\documentclass[review]{elsarticle} 
\usepackage{makecell}
\usepackage{indentfirst}
\usepackage{multirow}
\usepackage{graphicx}
\usepackage{natbib}
\usepackage{indentfirst}
\usepackage[table]{xcolor}
\usepackage{subfigure}
\usepackage{enumitem}
\usepackage{amsmath}
\usepackage{epstopdf}
\usepackage{booktabs,tabularx}
\usepackage{lineno,hyperref}
\usepackage{algorithm}
\usepackage{algorithmic}

\begin{document}

\title{An Enhanced Grey Wolf Optimizer with Elite Inheritance and Balance Search Mechanisms}

\author[*]{Jianhua Jiang}
\ead{jjh@jlufe.edu.cn}
\author[**]{Ziying Zhao}
\ead{ziying.zhao@autuni.ac.nz}
\author[**]{Weihua Li\corref{mycorrespondingauthor}}
\ead{weihua.li@aut.ac.nz}
\author[***]{Keqin Li}
\ead{lik@newpaltz.edu}
\address[*]{Jilin Province Key Laboratory of Fintech, Jilin University of Finance and Economics, Changchun 130117, P. R. China}
\address[**]{School of Engineering, Computer and Mathematical Sciences, Auckland University of Technology, Auckland, New Zealand}
\address[***]{Department of Computer Science, State University of New York, New Paltz, NY 12561, USA}

\cortext[mycorrespondingauthor]{Weihua Li is the corresponding author.}

\begin{abstract}
The Grey Wolf Optimizer (GWO) is recognized as a novel meta-heuristic algorithm inspired by the social leadership hierarchy and hunting mechanism of grey wolves. It is well-known for its simple parameter setting, fast convergence speed, and strong optimization capability. In the original GWO, there are two significant design flaws in its fundamental optimization mechanisms. \textbf{Problem (1):} the algorithm fails to inherit from elite positions from the last iteration when generating the next positions of the wolf population, potentially leading to suboptimal solutions. \textbf{Problem (2):} the positions of the population are updated based on the central position of the three leading wolves ($alpha$, $beta$, $delta$), without a balanced mechanism between local and global search. To tackle these problems, an enhanced \textbf{Grey Wolf Optimizer with Elite Inheritance Mechanism and Balance Search Mechanism}, named as \textbf{EBGWO}, is proposed to improve the effectiveness of the position updating and the quality of the convergence solutions. The IEEE CEC 2014 benchmark functions suite and a series of simulation tests are employed to evaluate the performance of the proposed algorithm. The simulation tests involve a comparative study between EBGWO, three GWO variants, GWO and two well-known meta-heuristic algorithms. The experimental results demonstrate that the proposed EBGWO algorithm outperforms other meta-heuristic algorithms in both accuracy and convergence speed. Three engineering optimization problems are adopted to prove its capability in processing real-world problems. The results indicate that the proposed EBGWO outperforms several popular algorithms.
\end{abstract}

\maketitle

\section{Introduction} 
Optimization plays a key role in problem-solving by trying to find the best solution from all possible options, while taking into account certain restrictions \citep{1}. This principle is widely used in real-world situations, including scientific research, engineering designs, production management, and socio-economic issues \citep{2}. As an improvement of heuristics, meta-heuristic algorithms are gradually taking the place of some classic optimization methods, such as the direct search and gradient search methods \citep{3}. They are especially good at dealing with optimization problems that have nonlinear objective functions and complex mathematical procedures \citep{4}. 

Among these, the Grey Wolf Optimizer (GWO) algorithm \citep{5} stands out due to its effective optimization ability and simplicity of use, including fewer parameters and lower computational cost \citep{6,7}. In the classical GWO algorithm, positions of three leading wolves ($\overrightarrow{X}_{\alpha}$, $\overrightarrow{X}_{\beta}$, $\overrightarrow{X}_{\delta}$) guide the entire population in hunting their prey ($\overrightarrow{X}_{p}$). The algorithm uses control parameters $\overrightarrow{a}$, $\overrightarrow{A}$, $\overrightarrow{C}$ to adjust the search movements of each wolf $i$ ($i$=1, 2, ..., $n$), and adopts the search strategy to updates the distance ($\overrightarrow{D}_{i}$) between the prey and a wolf, as well as the position ($\overrightarrow{X}_{i}$) of a wolf. 

However, the GWO algorithm, as one of the meta-heuristic algorithms, suffers from some disadvantages, such as local optimum stagnation, imbalance between exploration and exploitation, and premature convergence, in finding optimal solutions \citep{8,9}. Therefore, the improvement of the GWO algorithm to obtain more efficient optimal solutions and convergence, and balance the exploration and exploitation performance \citep{10,11,12,13,14} is a crucial task in further research.

\subsection{Motivations}\label{subsec:Motivations}
Numerous researchers have put considerable effort into improving the performance of the Grey Wolf Optimizer (GWO) algorithm. They have adopted various enhancement methods to boost the GWO's performance. Some scholars have tried to balance the exploration and exploitation performance of the algorithm by adding or redesigning control parameters in the original GWO algorithm \citep{10,14}. Others have worked to improve the convergence performance of the algorithm by introducing new theories or methods, such as chaos theory \citep{11} and opposition-based learning approach \citep{12}. Additionally, some researchers have developed new hybrid algorithms \citep{6,7} by combining the GWO algorithm with other efficient algorithms to enhance overall performance.

However, despite these advancements, two significant issues remain largely unaddressed. These problems are the focus of our research and will be tackled in this paper. 

\begin{itemize}
    \item \textbf{Problem (1)}: The lack of an elite inheritance mechanism affects the performance of the GWO algorithm. Specifically, in the original GWO algorithm and most of its variants, a new set of positions will be regenerated at each iteration. Due to the randomness of the position updating process, the optimization process cannot effectively draw from previous experience. Consequently, the algorithm cannot consistently deliver better solutions within a specified time frame.
    
    \item \textbf{Problem (2)}: The current position updating mechanism leads to insufficient exploration capability of the basic GWO algorithm, increasing the risk of falling into the local optimum. Moreover, the hunting process design in the GWO algorithm causes a fast convergence speed, resulting in premature convergence and unbalanced exploitation and exploration performance. Therefore, it is crucial to strike an effective balance between exploration and exploitation.
\end{itemize}

\subsection{Contributions}\label{subsec:contributions}
Motivated by the challenges outlined, we present in this paper a novel variant of the Grey Wolf Optimizer (GWO), termed Enhanced Balance Grey Wolf Optimizer (EBGWO). This novel algorithm incorporates two key mechanisms: the \textbf{Elite Inheritance Mechanism} and the \textbf{Balance Search Mechanism}. 

\textbf{The elite inheritance mechanism} addresses the issue where all positions updated in one updating process fail to obtain superior fitness values and deviate from the optimal solution, a crucial problem in GWO. This mechanism, which has demonstrated effectiveness in numerous classical swarm intelligence algorithms such as the Tree Seed Algorithm \citep{1} and Genetic Algorithm \citep{15}, refines the position renewal process in each iteration. It selectively employs elite individuals from the previous positions to guide the updating of positions in the next iteration.

\textbf{The balance search mechanism} assists the wolves in escaping local optima and enhancing solution quality. The traditional GWO algorithm's position updating mechanism, which relies on the center positions of three candidate wolves, is a locally greedy optimization strategy. This strategy favors exploitation over exploration, potentially leading to inaccurate calculation of the optimal solution and trapping the algorithm in local optima. To address this, EBGWO introduces the Search Tendency ($ST$) operator, which excels in balancing exploration and exploitation. The algorithm divides the position updating process into two stages, dynamically adjusting between global and local search. In the first stage, the EBGWO algorithm redefines the candidate wolves using $\alpha$ wolf, $\beta$ wolf and a random wolf in the population to obtain the final position. This strategy promotes diversity and encourages exploration of the search space. In the second stage, the original position updating mechanism of the GWO algorithm is adopted due to its strong exploitation capability. This allows the algorithm to intensively search locally, thereby improving the precision of the optimal solution.

The performance of the proposed EBGWO algorithm is evaluated in subsequent sections of this paper. This evaluation involves benchmark function testing experiments, statistical testing, and real-world engineering design experiments. The experimental results reveal that the EBGWO algorithm demonstrates superior convergence effects and consistently obtains better solutions. Notably, compared to other algorithms, the proposed EBGWO algorithm exhibits a stronger ability to avoid local optima and keeps a better balance between exploration and exploitation.

To summarize, the \textbf{contributions} of this paper are listed as follows.

\begin{itemize}
    \item We proposed \textbf{the elite inheritance mechanism} to improve the quality of the solution and increase the convergence speed. 
    \item We used \textbf{the balance search mechanism} to modify the original position updating method, which can enhance the exploration capability of the algorithm and improve the performance of generating new positions in the next iteration.
    \item We analyzed the results of comparative experiments and engineering design optimization problems. The results show that the EBGWO algorithm outperforms the other excellent variants and algorithms in local optimal stagnation avoidance and the balance capabilities between exploration and exploitation, etc. In addition, the EBGWO algorithm obtains superior solutions in engineering design problems.
\end{itemize} 

\section{Related work}\label{sec:related work}
This section first introduces the source of inspiration and operating mechanism of the Grey Wolf Optimizer (GWO). Next, it reviews the related studies on the GWO algorithm in recent years, including the different types of variants that have emerged. Finally, it summarizes the limitations of existing works.

\subsection{An overview of GWO}\label{subsec:GWO}
The GWO algorithm is a popular swarm-intelligence algorithm introduced by Mirjalili \citep{5}. It has been cited approximately 6000 times according to the Web of Science database. The GWO algorithm and its variants are widely used in feature selection, image segmentation, neural networks, scheduling problems, etc. \citep{16,17,18,19}.

The GWO algorithm is inspired by the behaviour of grey wolves in nature, and it imitates the social leadership hierarchy and hunting mechanism. The population of wolves is divided into four groups based on the strict leadership hierarchy, which involves alpha ($\alpha$), beta ($\beta$), delta ($\delta$), and omega ($\omega$). The optimal candidate with the best fitness value is $ \alpha $, followed by $ \beta $ and $ \delta $. The remaining wolves update their positions in each iteration following the leading wolves.

The hunting mechanism of the GWO algorithm mainly includes three actions: tracking, encircling and attacking. The leading wolves $\alpha$, $\beta$, and $\delta$ in the wolf pack are responsible for guiding the direction of the entire population. The positions of $\omega$ wolves are updated according to the position vectors of the three leading wolves, and the final solution will converge in the space defined by the leading wolves. The hunting mechanism is implemented as follows.

The grey wolves encircle the prey by Eqs. (\ref{1}) and (\ref{2}).
\begin{equation}
\label{1}
\overrightarrow{D}=|\overrightarrow{C} \times \overrightarrow{X}_{p}(t)-\overrightarrow{X}(t)|.
\end{equation}
\begin{equation}
\label{2}
\overrightarrow{X}(t+1)=\overrightarrow{X}_{p}(t)-\overrightarrow{A} \times \overrightarrow{D}.
\end{equation}
where \emph{$t$} indicates the current iteration number, \emph{$\overrightarrow{A}$} and \emph{$\overrightarrow{C}$} are coefficient vectors, \emph{$\overrightarrow{X}_{p}$} is the position vector of the prey, \emph{$\overrightarrow{X}$} refers to the position vector of a grey wolf, and \emph{$\overrightarrow{D}$} is the distance between the prey and the wolf.

The controlling parameters in the algorithm involve $ \overrightarrow{A} $, $ \overrightarrow{C} $ and $ \overrightarrow{a} $, which are calculated in Eqs. (\ref{3}) to (\ref{6}).
\begin{equation}
\label{3} 
\overrightarrow{A}_{1}=2\overrightarrow{a} \times \overrightarrow{r}_{1}-\overrightarrow{a}, \overrightarrow{C}_{1}=2\overrightarrow{r}_{2}.
\end{equation}
\begin{equation}
\label{4} 
\overrightarrow{A}_{2}=2\overrightarrow{a} \times \overrightarrow{r}_{1}-\overrightarrow{a}, \overrightarrow{C}_{2}=2\overrightarrow{r}_{2}.
\end{equation}
\begin{equation}
\label{5} 
\overrightarrow{A}_{3}=2\overrightarrow{a} \times \overrightarrow{r}_{1}-\overrightarrow{a}, \overrightarrow{C}_{3}=2\overrightarrow{r}_{2}.
\end{equation}
\begin{equation}
\label{6}
\overrightarrow{a}=2-t\times(\dfrac{2}{T}).
\end{equation}
where $\overrightarrow{A}_{1}$, $ \overrightarrow{A}_{2}$, $\overrightarrow{A}_{3}$, $\overrightarrow{C}_{1}$, $ \overrightarrow{C}_{2}$ and $\overrightarrow{C}_{3}$ are used to control the movement of three leading wolves respectively. \emph{$\overrightarrow{a}$} is linearly decreased from 2 to 0 over iterations, \emph{$ T $} represents the maximum number of iterations, \emph{$\overrightarrow{r}_{1}$} and \emph{$\overrightarrow{r}_{2}$} are randomly selected vectors in the interval [0, 1].

$\overrightarrow{X}_{1}, \overrightarrow{X}_{2}$ and $\overrightarrow{X}_{3}$ are wolves generated around the three leading wolves $ \alpha $, $ \beta $, and $ \delta $. The search wolves ($\omega$) update their positions based on the Eq. (\ref{13}). The equations are Eqs. (\ref{7}) to (\ref{13}).
\begin{equation}
\label{7} 
\overrightarrow{D}_{\alpha}=|\overrightarrow{C}_{1}\times \overrightarrow{X}_{\alpha}-\overrightarrow{X}(t)|.
\end{equation}
\begin{equation}
\label{8} 
\overrightarrow{D}_{\beta}=|\overrightarrow{C}_{2}\times \overrightarrow{X}_{\beta}-\overrightarrow{X}(t)|.
\end{equation}\begin{equation}
\label{9} 
\overrightarrow{D}_{\delta}=|\overrightarrow{C}_{3}\times \overrightarrow{X}_{\delta}-\overrightarrow{X}(t)|.
\end{equation}\begin{equation}
\label{10} 
\overrightarrow{X}_{1}=\overrightarrow{X}_{\alpha}-\overrightarrow{A}_{1}\times(\overrightarrow{D}_{\alpha}).
\end{equation}\begin{equation}
\label{11} 
\overrightarrow{X}_{2}=\overrightarrow{X}_{\beta}-\overrightarrow{A}_{2}\times(\overrightarrow{D}_{\beta}).
\end{equation}\begin{equation}
\label{12} 
\overrightarrow{X}_{3}=\overrightarrow{X}_{\delta}-\overrightarrow{A}_{3}\times(\overrightarrow{D}_{\delta}).
\end{equation}\begin{equation}
\label{13} 
\overrightarrow{X}(t+1)=\dfrac{(\overrightarrow{X}_{1}+\overrightarrow{X}_{2}+\overrightarrow{X}_{3})}{3}.
\end{equation}

\begin{figure}[htpb]
\centering
\includegraphics[width=4.5in]{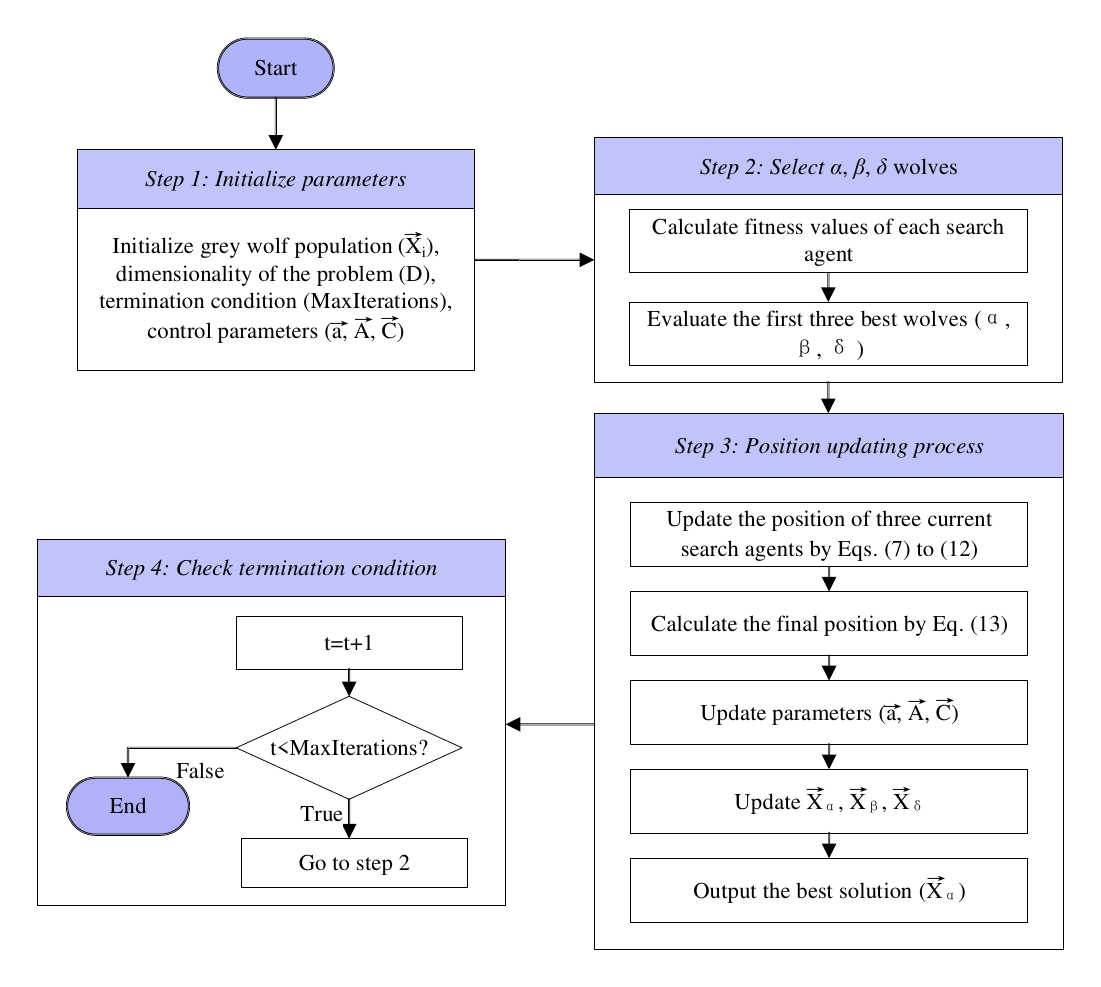} 
\caption{Flow chart of the Grey Wolf Optimization algorithm}
\label{fig.2}
\end{figure}

The pseudo-code of the basic GWO algorithm and the flow chart of the algorithm can be found in Algorithm \ref{alg:GWO} and Fig. \ref{fig.2}, respectively.

\begin{algorithm}[htbp]
\caption{: The Basic Grey Wolf Optimization Algorithm}         
\label{alg:GWO}                           
\begin{algorithmic}[1]
\REQUIRE \textit{$\overrightarrow{X_{i}}$, $Dim$, $MaxIterations$, $\overrightarrow{a}$, $\overrightarrow{A}$, $\overrightarrow{C}$}   
\ENSURE \textit{$ \overrightarrow{X_{\alpha}} $}     
\STATE Initialize the grey wolf population (\textit{$\overrightarrow{X_{i}}$}) \textit{(i=1, 2,$\cdots$, n)}
\STATE  Initialize the control parameters \textit{$ \overrightarrow{a} $}, \textit{$ \overrightarrow{A} $}, and \textit{$ \overrightarrow{C} $} 
\WHILE {\textit{$ t $} $<$ \textit{$MaxIterations$}}
\STATE  Calculate the fitness of each search wolf
\STATE  Evaluate the first three best wolves
\STATE  $ \overrightarrow{X_{\alpha}} $=the best search wolf
\STATE  $ \overrightarrow{X_{\beta}}$=the second best search wolf
\STATE  $ \overrightarrow{X_{\delta}}$=the third best search wolf
\FOR {each search wolf}
\STATE  Update positions of three current search wolves by Eqs. (7) to (12)
\STATE  Calculate the final position by Eq. (13)
\ENDFOR
\STATE  Update parameters \textit{$ \overrightarrow{a} $}, \textit{$ \overrightarrow{A} $}, and \textit{$ \overrightarrow{C} $}
\STATE  Update $ \overrightarrow{X_{\alpha}}$, $  \overrightarrow{X_{\beta}} $, and $  \overrightarrow{X_{\delta}}$
\STATE  $ t $=$ t $+1
\ENDWHILE
\STATE  \textbf{return $ \overrightarrow{X_{\alpha}}$} 
\end{algorithmic}
\end{algorithm}	

The attacking and searching mechanisms mean that the algorithm achieves the balance of exploration and exploitation by controlling the behaviour of individuals in the population. In the GWO algorithm, two coefficient vectors, i.e., $ \overrightarrow{A} $ and $ \overrightarrow{C} $ are adopted to control the movement of wolves. 

However, it can be found that the current attacking and searching mechanism with $ \overrightarrow{A} $ and $ \overrightarrow{C} $ vectors is not efficient enough to keep a good balance between local search and global search. The researchers need to design a more useful mechanism for the balance between exploration and exploitation.

\subsection{GWO variants}\label{subsec:VariantsGWO}
GWO has several notable \textbf{advantages}, including simple parameter settings, fast convergence speed, and easy implementation \citep{7,10,11,12,13,14,20}. However, the basic GWO algorithm has some shortcomings, such as an imbalance between exploration and exploitation, local optimum, etc. Various GWO variants have been designed to alleviate these problems \citep{21}.

\textbf{New operators modification:} Long et al. \citep{10} add a nonlinear modulation index $\mu$ to adapt the parameter of \textit{$ a $}, providing a better balance the exploration and exploitation. Kohli et al. \citep{11} introduce chaos theory into the GWO algorithm to expedite global convergence. Yu et al. \citep{12} modify the $\overrightarrow{a}$ value using a nonlinear function and add a jumping rate based on the opposition-based learning approach to improve the performance. Mittal et al. \citep{14} use the exponential decay function to control the decrease of \textit{$\overrightarrow{a}$}, reallocating the number of iterations for exploration and exploitation. These works modify the algorithm by adding new operators or modifying the parameter $\overrightarrow{a}$ in the original GWO rather than changing the mechanism of the algorithm.

\textbf{Encoding scheme of the individuals modification:} Dhargupta et al. \citep{13} combine the selective opposition theory with the original GWO to improve exploration capability. Liu et al. \citep{22} design a hybrid model, namely a binary grey wolf echo state network, to improve the network's generalization capabilities. Zhu et al. \citep{19} propose a discrete variant of GWO and combine it with Cellular automata (CA). The shuffled cellular evolutionary grey wolf optimizer (SCEGWO) performs well in solving flexible job shop scheduling problems with job precedence constraints (FJSSP-JPC).

\textbf{Position updating mechanism modification:} Meng et al. \citep{18} introduce a new variant named AGWO to modify a new location updating strategy. The authors use elastic, circling, and attacking mechanisms to alter the updating mechanism. Gupta et al. \citep{20} propose a new variant of GWO named random walk grey wolf optimizer (RW-GWO) based on the random walk process to improve GWO exploration. Jiang et al. \citep{23} provide an improved GWO algorithm based on diversity enhanced strategy (DSGWO) by changing the number of leading search wolves and the population updating process. These algorithms improve the position updating mechanism by introducing a new theory or new evaluation methods. In this regard, we make a new design, i.e., \textbf{the balance search mechanism}.

\textbf{Population structure and hierarchy modification:} Yang et al. \citep{24} adopt a cooperative hunting group and a random scout group to balance the exploration and exploitation capabilities. Saremi et al. \citep{25} apply evolutionary population dynamics operators into GWO, improving the overall median fitness and increasing the convergence speed. Wu et al. \citep{42} provide a multi-population topology strategy to improve the population diversity and an information interaction learning strategy is adopted to change the position updating mechanism. Their proposed MTBGWO algorithm outperforms some recent algorithms in solving feature selection problems. MTBGWO algorithm is designed for multi-objective combinatorial optimization, while our proposed EBGWO algorithm is for single-objective continuous problems, each using unique strategies for optimization. We propose the \textbf{elite inheritance mechanism} to improve the population renewal process. The mechanism combines elite leading wolves in the last iteration and the current iteration and designs a selection of leading wolves in the candidate pool to guide the position updating process.

Tables \ref{relatedwork-table1}, \ref{relatedwork-table2}, and \ref{relatedwork-table3} present a summary of related work in terms of new design mechanisms, data sets, and evaluation metrics. These methods mainly focus on the modification of the current iteration positions or the position updating mechanism, but there are few research works that try to change the population positions renewal mechanism. Different from the above variants, the EBGWO algorithm proposed in this paper improves GWO in three aspects: introducing a new operator, changing the structure and renewal process of the population and modifying the position updating process in the next iteration. 

\begin{table}[ht]
  \centering
  \caption{Some variants of GWO algorithm in terms of new design mechanisms}
    \begin{tabular}{lc}
    \hline
    \multirow{2}[0]{*}{Algorithms} & Criteria\\
    \cline{2-2} 
    \multicolumn{1}{c}{} & New design mechanisms \\\hline
    
    MAL-IGWO (2017) & \makecell[c]{ New operators modification }\\
    \cline{2-2} CGWO (2017) & New operators modification \\
   \cline{2-2}  OGWO (2021) & New operators modification \\
    \cline{2-2} mGWO (2016) & New operators modification \\
    \cline{2-2} SOGWO (2020) &\makecell[c]{Encoding scheme of the \\ individuals modification }\\
    \cline{2-2} BGWO-ESN (2020) & \makecell[c]{Encoding scheme of the \\ individuals modification \& Binary }\\
    \cline{2-2} SCEGWO (2022) & \makecell[c]{Encoding scheme of the individuals\\ modification \& Population structure \\ and hierarchy modification} \\
    \cline{2-2} AGWO (2021) & Position updating mechanism modification \\
   \cline{2-2}  RW-GWO (2019) & Position updating mechanism modification \\
   \cline{2-2}  DSGWO (2022) & Position updating mechanism modification \\
   \cline{2-2}  GGWO (2017) & Population structure and hierarchy modification \\
   \cline{2-2}  GWO-EPD (2015) & Population structure and hierarchy modification \\
   \cline{2-2}  MTBGWO (2023) & \makecell[c]{Position updating mechanism modification \&\\ Population structure and hierarchy modification \& \\ Binary} \\\hline
    \end{tabular}
  \label{relatedwork-table1}%
\end{table}%

\begin{table}[ht]
  \centering
  \caption{Some variants of GWO algorithm in terms of data sets}
    \begin{tabular}{lc}
    \hline
    \multirow{2}[0]{*}{Algorithms} & Criteria \\
    \cline{2-2} 
    \multicolumn{1}{l}{} & Data sets \\\hline
    MAL-IGWO (2017) & \makecell[c]{CEC 2006 benchmark functions \& \\ 3 engineering design optimization problems} \\
   \cline{2-2}  CGWO (2017) & \makecell[c]{13 constrained benchmark functions \& \\ 5 engineering design optimization problems} \\
    \cline{2-2} OGWO (2021) & 23 benchmark functions \\
    \cline{2-2} mGWO (2016) & \makecell[c]{27 benchmark functions \& \\ 1 clustering problem in WSN}  \\
    \cline{2-2} SOGWO (2020) & 23 benchmark functions \\
   \cline{2-2}  BGWO-ESN (2020) & \makecell[c]{benchmarking database for time series prediction \& \\ real-world financial data sets} \\
    \cline{2-2} SCEGWO (2022) & \makecell[c]{15 different scaled FJSSP-JPC instances \& \\ a case study on the multi-category \\ bicycle production scheduling} \\
   \cline{2-2}  AGWO (2021) & \makecell[c]{CEC 2014 benchmark functions \& \\ 7 classification data sets \& \\ 3 function approximation data sets} \\
    \cline{2-2} RW-GWO (2019) & \makecell[c]{ CEC 2014 benchmark functions \& \\ 4 real life application problems } \\
   \cline{2-2}  DSGWO (2022) & \makecell[c]{CEC 2014 benchmark functions \& \\ 2 engineering design optimization problems} \\
   \cline{2-2}  GGWO (2017) & case study \\
   \cline{2-2}  GWO-EPD (2015) & 13 benchmark functions \\
   \cline{2-2}  MTBGWO (2023) & \makecell[c]{16 data sets from the UCI database }  \\\hline
    \end{tabular}%
  \label{relatedwork-table2}%
\end{table}%

\begin{tiny}
\begin{table}[ht]
  \centering
  \caption{\label{relatedwork-table3} Some variants of GWO algorithm in terms of evaluation metrics}
\begin{tabular}{lcc}
    \hline
    \multirow{2}[0]{*}{Algorithms} & Criteria \\
    \cline{2-2} 
    \multicolumn{1}{l}{} & Evaluation metrics \\\hline
    
    MAL-IGWO (2017) & Best, Mean, Average, Worst, Std \\
    \cline{2-2}CGWO (2017) & \makecell[c]{Best, Mean, Worst, Std, \\ Wilcoxon signed rank-test} \\
    \cline{2-2}OGWO (2021) & \makecell[c]{Mean, Std, Wilcoxon signed rank-test} \\
   \cline{2-2} mGWO (2016) & Average, Std \\
    \cline{2-2}SOGWO (2020) & \makecell[c]{Best, Average, Std, Friedman Test,\\ Kruskal-Walli’s test, post-hoc multiple comparison test} \\
    \cline{2-2}BGWO-ESN (2020) & \makecell[c]{NMSE, Accuracy improvement, Running time \& \\ RMSE, Accuracy improvement} \\
   \cline{2-2} SCEGWO (2022) & Friedman test, Wilcoxon signed rank test \\
    \cline{2-2}AGWO (2021) & \makecell[c]{Mean, Std, Worst, Best, Median \&\\ MSE, Std, Accuracy/ Error} \\
    \cline{2-2}RW-GWO (2019) & \makecell[c]{Mean, Std Dev., Median, Minimum,\\ Maximum, Wilcoxon signed rank-test, \\ Performance Index (PI)} \\
    \cline{2-2}DSGWO (2022) & \makecell[c]{Mean, Std, (w/t/l) Ranking,\\ Overall Effectiveness, Wilcoxon signed rank-test} \\
   \cline{2-2} GGWO (2017) & \makecell[c]{Time, System responses, Maximization, \\ Minimization, Mean, Worst, Best, \\ Std. Dev., Rel. Std. Dev. }\\
    \cline{2-2}GWO-EPD (2015) & Maximization, Minimization \\
    \cline{2-2}MTBGWO (2023) & \makecell[c]{Average Accuracy, Mean fitness, Best fitness, \\ Worst fitness, Average selection, Average time,\\ Wilcoxon signed rank test} \\\hline
    \end{tabular}
\end{table}
\end{tiny}

\section{The EBGWO algorithm}\label{sec:method}
As far as the current state of research is concerned, the convergence effect and the balance of exploration and exploitation of the GWO algorithm still require further improvement. Therefore, we propose a novel GWO variant, which incorporates the elite inheritance and balance search mechanisms. The proposed EBGWO algorithm is capable of producing more promising results than other popular GWO variants. The structure of EBGWO and its corresponding computational complexity are elaborated in the following subsections.

\subsection{Elite inheritance mechanism}
The proposed elite inheritance mechanism builds an \textbf{Elite Archive} to enhance the convergence effect of the EBGWO algorithm. 

\begin{figure}[htbp]
\centering
\includegraphics[width=5in]{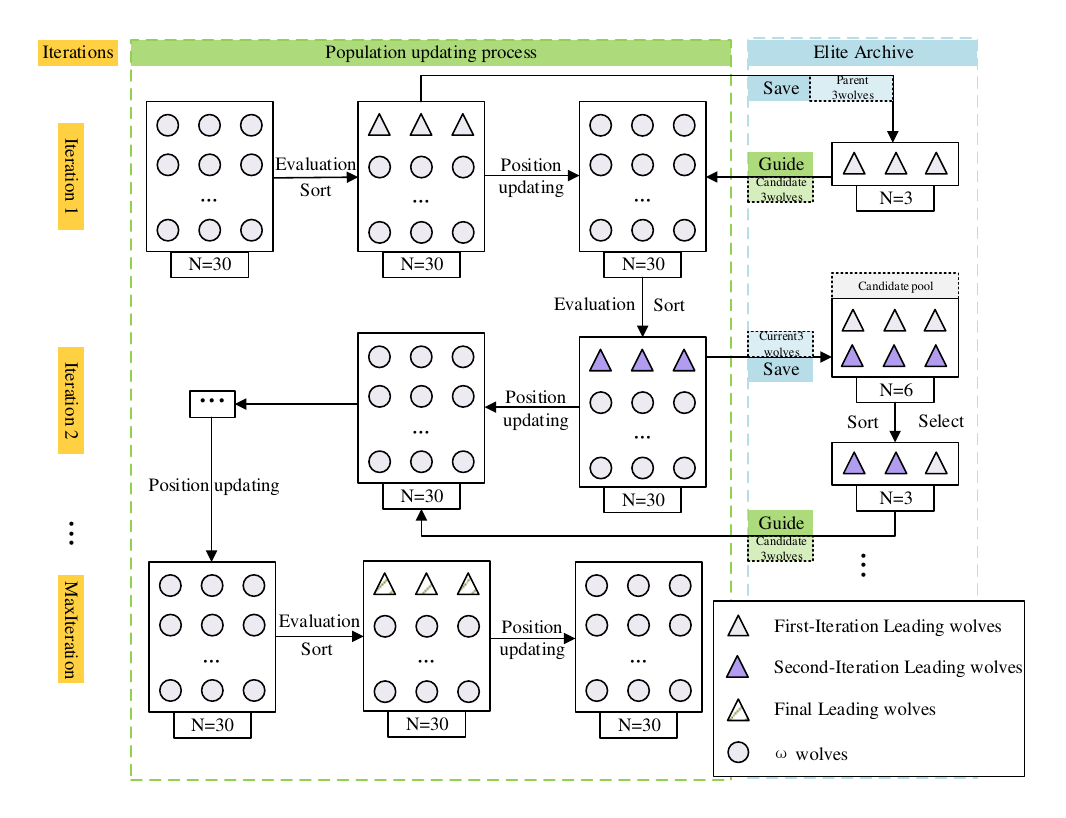}
\caption{The elite inheritance mechanism in EBGWO}
\label{fig.3} 
\end{figure}

The elite inheritance mechanism is shown in Fig. \ref{fig.3}, where a population of size 30 is utilized. In the first iteration, the initial population is generated and sorted according to their evaluation fitness values. Next, the first three leading wolves with the best fitness values are selected as the $\alpha$ position, $\beta$ position, and $\delta$ position. These positions are saved in the group named \textbf{Parent3wolves}. Then, the positions of the entire population are updated according to the \textbf{Candidate3wolves} (which in the $1^{st}$ iteration is equal to \textbf{Parent3wolves}) in the \textbf{Elite Archive}. In the next iteration, the \textbf{Candidate3wolves} from the last iteration is inherited and saved in the \textbf{Parent3wolves}. Moreover, the current positions of three leading wolves are stored in the \textbf{Current3wolves}. Then, the \textbf{Parent3wolves} and the \textbf{Current3wolves} are combined to form the \textbf{Candidate pool}. Finally, the algorithm sorts the \textbf{Candidate pool} and selects the new three leading wolves as the \textbf{Candidate3wolves} to guide the position updating process. Thus, the elite positions generated in each iteration are inherited and adopted, improving the quality of solutions. In addition, the application of this strategy implies that the produced next iteration positions will not be entirely inferior to the previous positions. 

\begin{figure}[ht]
\centering
\includegraphics[width=4.8in]{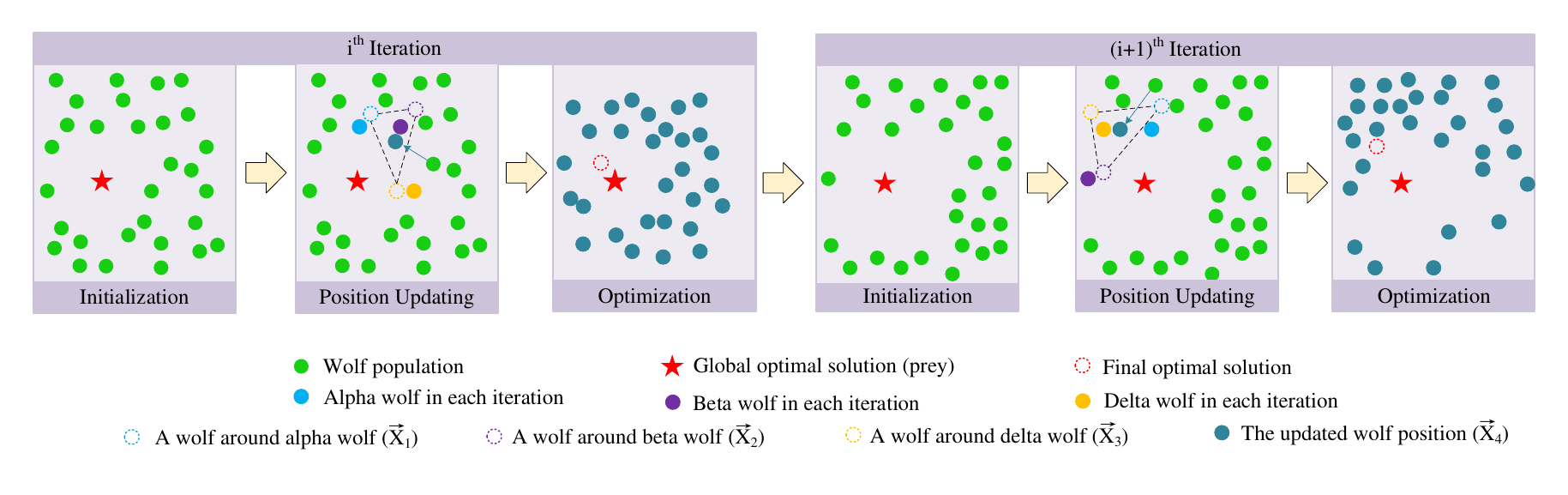}
\caption{The optimization process of two iterations ($i^{th}$ and $(i+1)^{th}$) in GWO}
\label{fig.GWO} 
\end{figure}

The proposed elite inheritance mechanism is inspired by GA \citep{15}. In GA, after the initial population is evaluated, individuals with better fitness values are retained for the subsequent iteration, while those with poor fitness values are eliminated. The population updating process of the original GWO is significantly different from GA. Unlike GA, GWO aims to identify the top three wolves in the $i^{th}$ iteration after evaluation and updates the entire population locations based on the three wolves' positions. As depicted in Fig. \ref{fig.GWO}, the GWO algorithm does not have the elite archive to store the elite positions in the last iteration. Instead, it relies on the three current leading wolves for hunting. If the optimal solution ($\alpha$ wolf) in the $i^{th}$ iteration is closer to the prey than in the $(i+1)^{th}$ iteration, it is possible that the fitness value of the optimal solution may not improve from the previous iteration. This suggests that the solution found in the next iteration may deviate from the global optimal solution, leading to premature convergence and a local optimum. Unlike GA, the elite inheritance mechanism proposed in this paper does not employ crossover and mutation operators. To address the limitations mentioned above, this paper adopts a new selection method. The pseudo-code for \textbf{the elite inheritance mechanism} is presented in Algorithm \ref{alg:1mechanism}. 

\begin{algorithm}[htbp]  
\caption{: Elite inheritance mechanism in EBGWO}
\label{alg:1mechanism}                                              
\begin{algorithmic}[1]
\REQUIRE \textbf{$\overrightarrow{X_{\alpha}}$}, \textbf{$\overrightarrow{X_{\beta}}$}, \textbf{$\overrightarrow{X_{\delta}}$}, fitness values \textbf{$f_{i}$ $(i=1,$ $\cdots$, $n)$}
\ENSURE \textit{candidate3wolves}
\IF {\textit{$ t $} $==$ 0}
\STATE parent3wolves $\leftarrow$ [\textit{$\overrightarrow{X_{1}},$ $\cdots$, $\overrightarrow{X_{k}}$}]; // $(k=1, 2, 3)$ \COMMENT{1: $\alpha$, 2: $\beta$, 3: $\delta$}
\STATE candidate3wolves $\leftarrow$ parent3wolves;
\ELSE
\STATE parent3wolves $\leftarrow$ candidate3wolves;
\STATE current3wolves $\leftarrow$ [\textit{$\overrightarrow{X_{1}}$, $\cdots$, $\overrightarrow{X_{j}}$}]; // $(j=1, 2, 3)$ \COMMENT{1: $\alpha$, 2: $\beta$, 3: $\delta$}
\STATE candidate pool $\leftarrow$ [parent3wolves current3wolves];
\STATE $sort_{m}$ $\lbrace$ $f_{m}$ $\rbrace$ in candidate pool; // $(m=k+j$, $m=1, 2,$ $\cdots$,$ 6)$
\STATE candidate3wolves $\leftarrow$ candidate pool [\textit{$\overrightarrow{X_{1}},$ $\cdots$, $\overrightarrow{X_{n}}$}]; // (n=1, 2, 3)
\ENDIF
\STATE \textbf{return} \textit{candidate3wolves}
\end{algorithmic}
\end{algorithm}

\subsection{Balance search mechanism}
In the original GWO algorithm, the position updating mechanism tends to fall into the local optimum \citep{18,20}. To address this problem, the balance search mechanism is proposed to improve the position updating process. The EBGWO algorithm introduces a \textbf{new operator}, $ST$, to segment the updating process and \textbf{redefines leading wolves} to broaden the solution space. 

\begin{figure}[!t]
\centering
\includegraphics[width=4.8in]{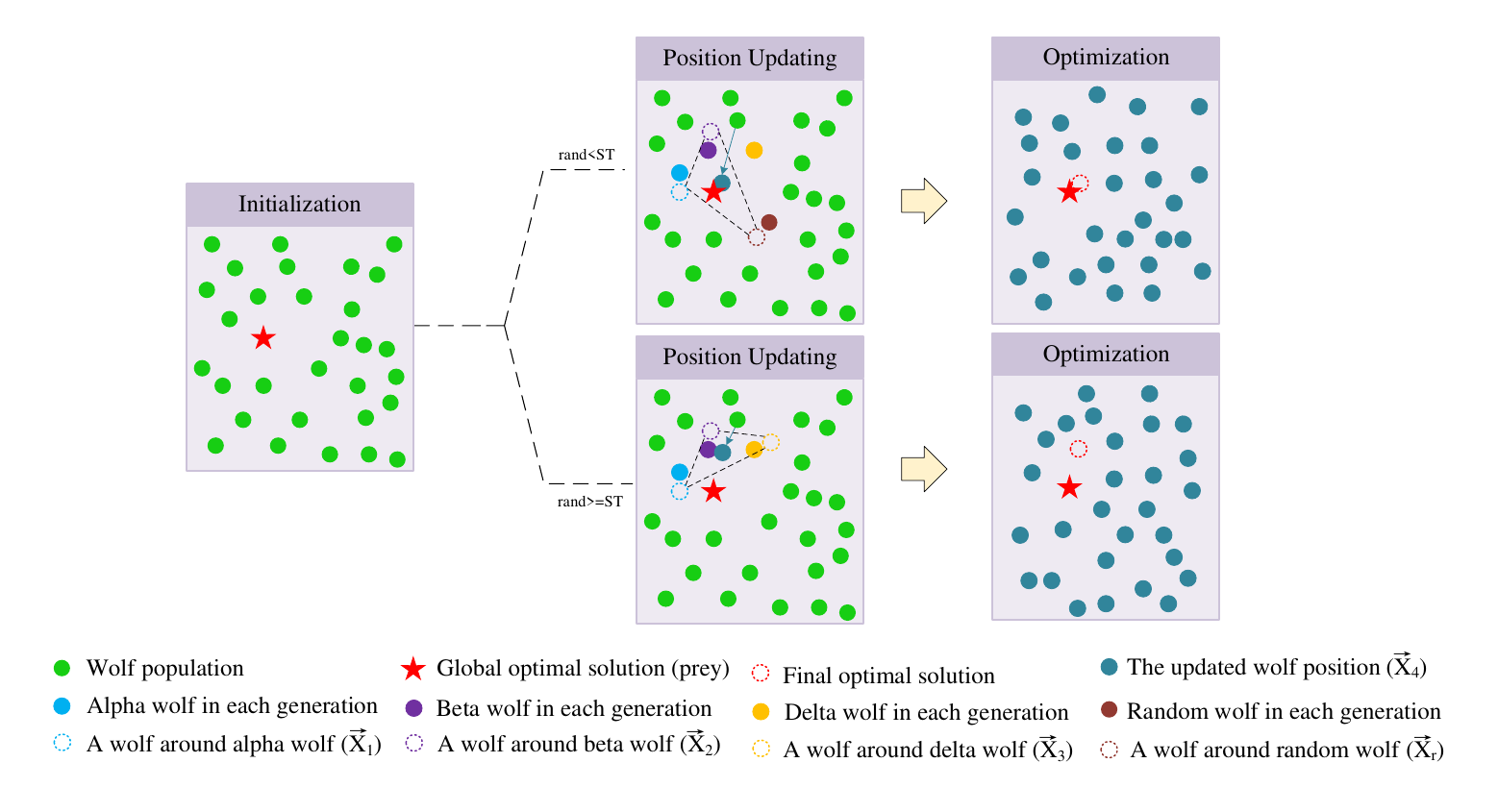}
\caption{Balance search mechanism of EBGWO}
\label{fig.ST} 
\end{figure}

As shown in Fig. \ref{fig.ST}, the proposed balance search mechanism is mainly implemented through two components. The selection of the position update equations is controlled by the $ST$ parameter. When the $rand$ value (a random value within the range [0, 1]) is less than the $ST$ value, the population explores the search region using Eqs. (\ref{14}) to (\ref{20}). Otherwise,  if the $rand$ value exceeds the $ST$ value, the position updating process will utilize the original GWO mechanism, as defined by Eqs. (\ref{7}) to (\ref{13}) to exploit the local search space. The new position updating mechanism, as defined by Eqs. (\ref{14}) to (\ref{20}), is realized through the redesign of the leading wolves. In Eqs. (\ref{14}) and (\ref{15}), the $\overrightarrow{X}_{candidate3wolves(1)}$ and $\overrightarrow{X}_{candidate3wolves(2)}$ are selected in the \textbf{candidate pool}, which are the new $\alpha$ and $\beta$. In Eq. (\ref{16}), the $\overrightarrow{X}_{r}$ is a random wolf in the population to extend the search area.

The proposed balance search mechanism capitalizes on the strengths of both the $ST$ operator and the newly leading wolves. The $ST$ operator, introduced in the TSA algorithm, balances exploration and exploitation within a biased distribution. In this paper, the $ST$ value is set to 0.2. In basic GWO, as shown in Eqs. (\ref{10}) to (\ref{12}), the $ \alpha $, $ \beta $ and $ \delta $ wolves update positions towards $ \overrightarrow{X}_{1} $, $ \overrightarrow{X}_{2} $ and $ \overrightarrow{X}_{3} $, respectively. In each iteration, the final position of $ \omega$ wolf to attack the prey is calculated according to the average value among the three leading wolves' positions in Eq. (\ref{13}). The position updating method constrains the solution to the central position of the three leading wolves, making the final solution more likely to be a local optimum rather than a global one. The balance search mechanism expands the selection range for the optimal solution, enables the algorithm to escape local optima, and ultimately enhances both the quality of the solution and the diversity of the global search. Leveraging the balance search mechanism, the improved optimizer can identify a precise global optimum.

\begin{equation}
\label{14} 
\overrightarrow{D}_{\alpha}=|\overrightarrow{C}_{1}\times \overrightarrow{X}_{candidate3wolves(1)}-\overrightarrow{X}(t)|.
\end{equation}
\begin{equation}
\label{15} 
\overrightarrow{D}_{\beta}=|\overrightarrow{C}_{2}\times \overrightarrow{X}_{candidate3wolves(2)}-\overrightarrow{X}(t)|.
\end{equation}
\begin{equation}
\label{16} 
\overrightarrow{D}_{\delta}=|\overrightarrow{C}_{3}\times \overrightarrow{X}_{r}-\overrightarrow{X}(t)|.
\end{equation}
\begin{equation}
\label{17} 
\overrightarrow{X}_{1}=\overrightarrow{X}_{candidate3wolves(1)}-\overrightarrow{A}_{1}\times(\overrightarrow{D}_{\alpha}).
\end{equation}
\begin{equation}
\label{18} 
\overrightarrow{X}_{2}=\overrightarrow{X}_{candidate3wolves(2)}-\overrightarrow{A}_{2}\times(\overrightarrow{D}_{\beta}).
\end{equation}
\begin{equation}
\label{19} 
\overrightarrow{X}_{3}=\overrightarrow{X}_{r}-\overrightarrow{A}_{3}\times(\overrightarrow{D}_{\delta}).
\end{equation}
\begin{equation}
\label{20} 
\overrightarrow{X}(t+1)=\dfrac{(\overrightarrow{X}_{1}+\overrightarrow{X}_{2}+\overrightarrow{X}_{3})}{3}.
\end{equation}

\subsection{The EBGWO algorithm}
The proposed EBGWO algorithm introduces the elite inheritance mechanism and the balance search mechanism to improve the performance further. 

In each iteration, the EBGWO algorithm identifies the leading wolves from the last iteration, stores them in \textbf{Parent3wolves}, selects the leading wolves from the current iteration into \textbf{Current3wolves}, and combines them in the \textbf{Candidate pool}. Three elite positions are identified in the \textbf{Candidate pool} and stored in \textbf{Candidate3wolves} to guide the position updating process. The $ST$ operator is adopted subsequently to balance the exploration and exploitation of the EBGWO algorithm. Lastly, the EBGWO algorithm either employs the newly defined leading wolves or applies the original GWO position updating mechanism to update the entire population.

In the basic GWO, a new population distribution is generated in each iteration, guided by the leading wolves from the previous iteration. Fig. \ref{fig.4}(a) illustrates the final positions of the whole population in a random place within a circle defined by the positions of $ \alpha $, $ \beta $, and $ \delta $ in the search space \citep{5}. As shown in Fig. \ref{fig.4}(b), the distance between $ \overrightarrow{X'}_{4} $ and the prey is shorter than that between $ \overrightarrow{X}_{4}$ and the prey. When the position of the prey deviates from the range defined by the three leading wolves, the solution may not be optimal. The next iteration position updating to $ \overrightarrow{X}_{4} $ is not always a good choice. In the EBGWO algorithm, a random $\omega$ wolf is selected as the new third leading wolf in the population, thereby expanding the search range and facilitating escape from local optima. Therefore, $ \overrightarrow{X'}_{4} $, utilizing the balance search mechanism, yields a better solution than $ \overrightarrow{X}_{4} $. A typical local optimal landscape example from CEC 2005 benchmark function suites can be found in Figure \ref{fig.11}. The multimodal benchmark function can be found in Table \ref{CEC2005} \citep{5}.

\begin{figure}[htbp]
\centering
\includegraphics[width=4.8in]{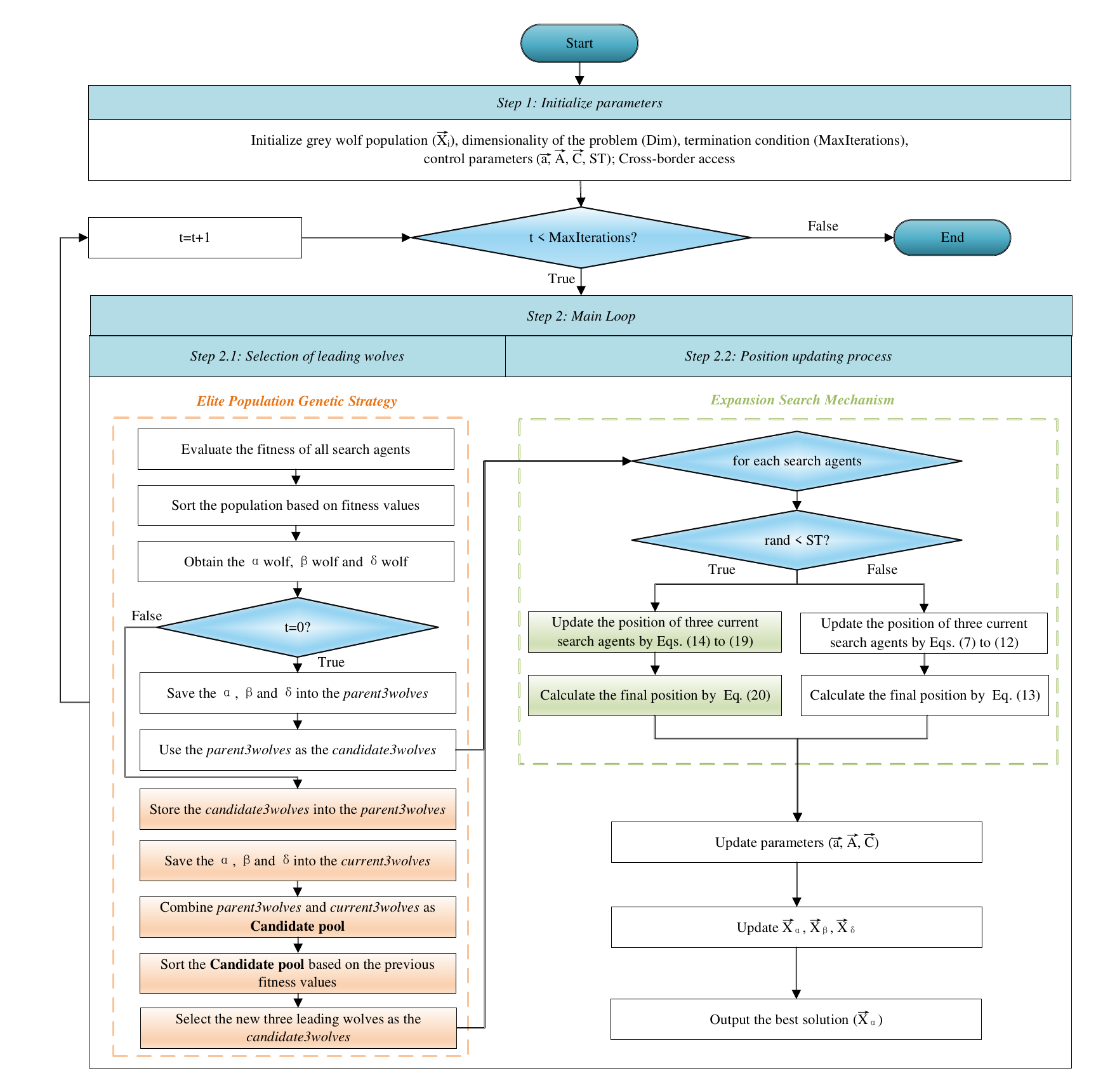}
\caption{Flow chart of EBGWO}
\label{fig.5} 
\end{figure}

\begin{figure}[htpb]
\centering
\includegraphics[width=4.8in]{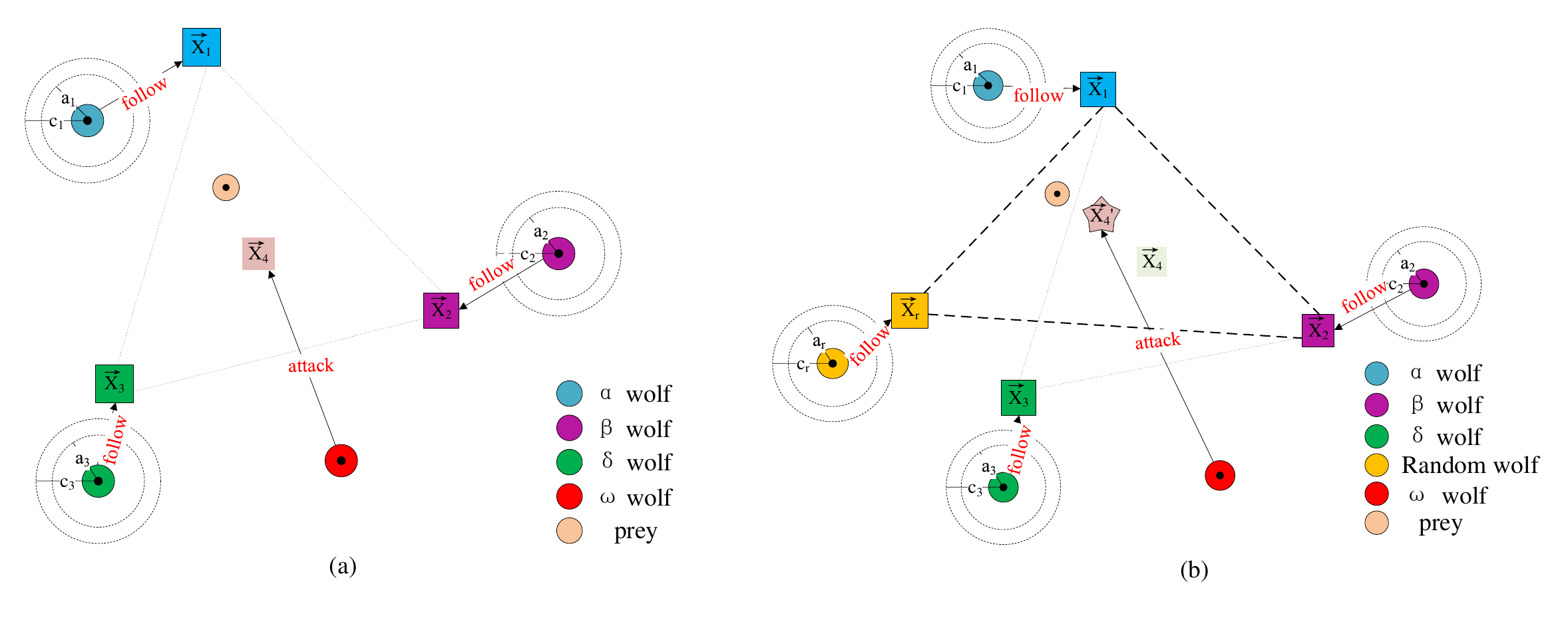}
\caption{Position updating process in GWO (a) and EBGWO (b)}
\label{fig.4}
\end{figure}

\begin{figure}[htpb]
\centering
\includegraphics[width=4.8in]{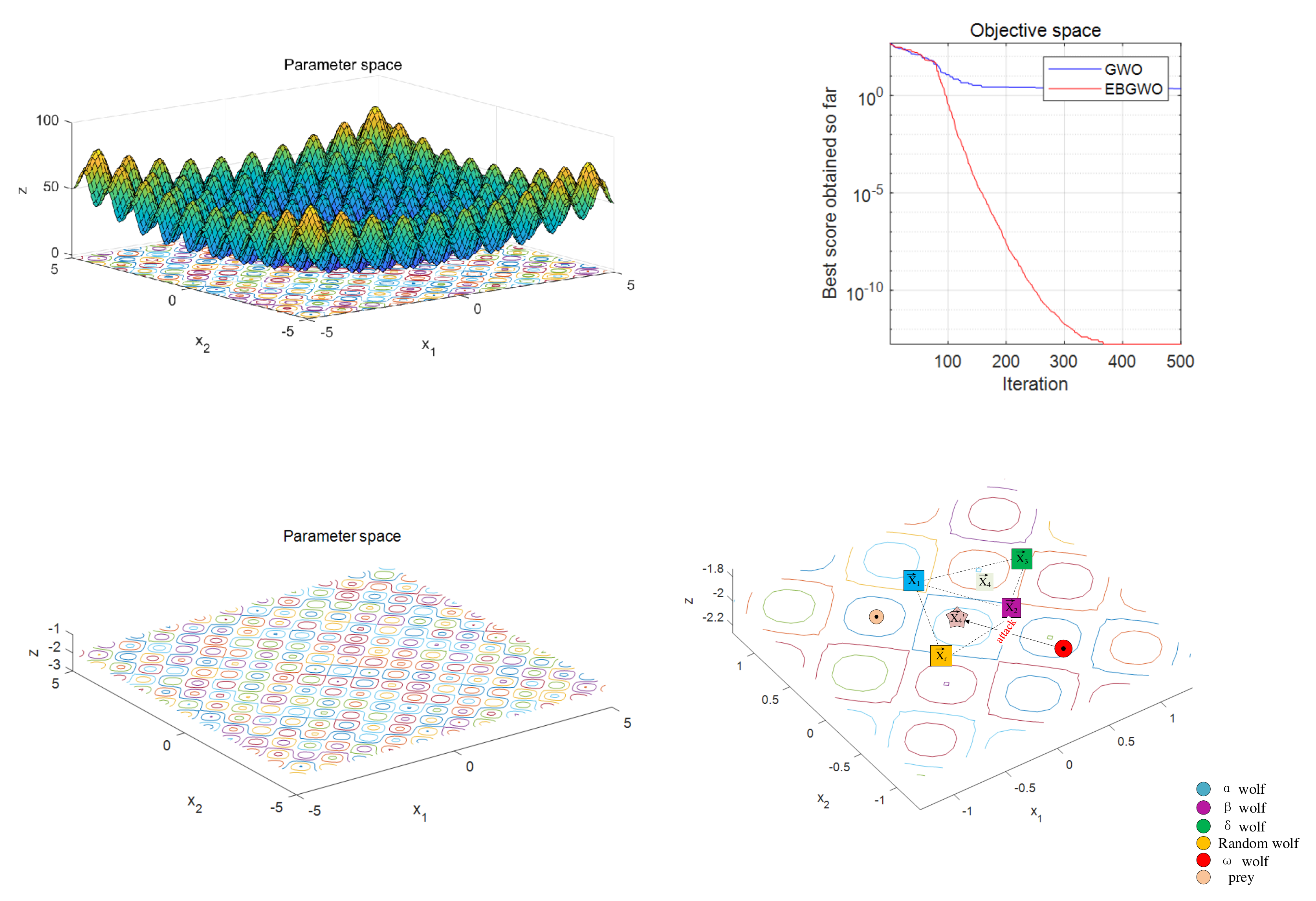}
\caption{Position updating process of EBGWO algorithm in typical local optimal landscape example of CEC 2005 benchmark functions (F9)}
\label{fig.11}
\end{figure}

\begin{table}[htbp]
\begin{center}
\caption{\label{CEC2005} Multimodal benchmark function $F9$ of IEEE CEC 2005}
\begin{tabular}{lccc}
\hline
Function & Dim  &  Range & $ f_{min}$\\
\hline
$F_9(x)=\Sigma^{n}_{i=1}[x_{i}^{2}-10cos(2\pi*x_{i}+10]$ & 30 & [-5.12, 5.12] &  0 \\
\hline
\end{tabular}
\end{center}
\end{table}
The pseudo-code of the proposed EBGWO algorithm is presented in Algorithm \ref{alg:EBGWO}, and the flow chart is presented in Fig. \ref{fig.5}.

\begin{algorithm}[htbp]                    
\caption{: EBGWO: The proposed Grey Wolf Optimization Algorithm}
\label{alg:EBGWO}                                              
\begin{algorithmic}[1]
\REQUIRE \textit{$\overrightarrow{X_{i}}$}, Dim, $MaxIterations$, $ \overrightarrow{a} $, $ \overrightarrow{A} $, $ \overrightarrow{C} $, $ST$ 
\ENSURE \textit{$\overrightarrow{X_{\alpha}}$}     
\STATE Initialize the grey wolf population (\textit{$\overrightarrow{X_{i}}$}) \textit{(i=1, 2, $\cdots$, n)}
\STATE Initialize the control parameter \textit{$ \overrightarrow{a} $}, \textit{$ \overrightarrow{A} $}, and \textit{$ \overrightarrow{C} $}, \textit{$ST$}=0.2 
\WHILE {\textit{$ t $} $<$ \textit{$MaxIterations$}}
\STATE Check the population whether it goes beyond the boundaries of the search space
\STATE Evaluate the fitness of all search wolves
\STATE Sort the population based on fitness values
\STATE Obtain the positions of $\alpha$ wolf, $\beta$ wolf and $\delta$ wolf
\STATE $\overrightarrow{X_{\alpha}} $=the best search wolf
\STATE $ \overrightarrow{X_{\beta}}$=the second best search wolf
\STATE $ \overrightarrow{X_{\delta}}$=the third best search wolf
\IF {\textit{$ t $} $==$ 0}
\STATE Save $\overrightarrow{X_{\alpha}}$, $\overrightarrow{X_{\beta}}$ and $\overrightarrow{X_{\delta}}$ into the \textit{parent3wolves}
\STATE Use the \textit{parent3wolves} as the \textit{candidate3wolves} of the position updating process
\ELSE
\STATE Store the \textit{candidate3wolves} into the \textit{parent3wolves}
\STATE Save $\overrightarrow{X_{\alpha}}$, $\overrightarrow{X_{\beta}}$ and $\overrightarrow{X_{\delta}}$ into the \textit{current3wolves}
\STATE Combine \textit{parent3wolves} and \textit{current3wolves} as a \textit{candidate pool}
\STATE Sort the \textit{candidate pool} based on the previous fitness values
\STATE Select the new three leading wolves as the \textit{candidate3wolves}
\ENDIF
\FOR {each search wolf}
\IF {\textit{$ rand $} $<$ \textit{$ST$}}
\STATE Update positions of three current search wolves by Eqs. (\ref{14}) to (\ref{19})
\STATE Calculate the final position by Eq. (\ref{20})
\ELSE
\STATE Update positions of three current search wolves by Eqs. (\ref{7}) to (\ref{12})
\STATE Calculate the final position by Eq. (\ref{13})
\ENDIF
\ENDFOR
\STATE Update \textit{$ \overrightarrow{a} $}, \textit{$ \overrightarrow{A} $}, and \textit{$ \overrightarrow{C} $}
\STATE Update $\overrightarrow{X_{\alpha}}$, $\overrightarrow{X_{\beta}}$ and $\overrightarrow{X_{\delta}}$
\STATE $ t $=$ t $+1
\ENDWHILE
\STATE \textbf{return \textit{$ \overrightarrow{X_{\alpha}}$}} 
\end{algorithmic}
\end{algorithm}	

\subsection{Computational complexity of EBGWO}
Utilizing the Big-O notation method \citep{26}, the paper establishes some indicators to evaluate the computational complexity of EBGWO and GWO. These include the population size ($n$), the iteration number ($t$), the cost of fitness function evaluation ($c$), and the dimension of the problem ($Dim$) \citep{23}. It is evident that the computational complexity of EBGWO and GWO is approximately the same. This similarity is due to the fact that the elite inheritance mechanism of EBGWO can reduce the computational cost of the position updating process and speed up the population search in the space. The CPU computational time for the EBGWO algorithm and GWO algorithm is presented in Tables \ref{time1} and \ref{time2}.

\begin{equation}
\label{21} 
O(GWO)=O(t \times Dim \times n^{2} + t \times Dim \times n \times c)
\end{equation}
\begin{equation}
\label{22} 
O(EBGWO)=O(t \times Dim \times n^{2}+t \times Dim \times n \times c)
\end{equation}

\begin{table}[htbp]
  \centering
  \caption{Computation time of EBGWO and GWO in 10D and 30D}
      \label{time1}
      \tabcolsep 7.3pt 
      \resizebox{3in}{43mm}{
    \begin{tabular}{c|cc|cc}
    \toprule
     \multicolumn{5}{c}{Computation Time} \\
    \hline
   \multirow{2}{*}{F}       &    \multicolumn{2}{c|}{D=10} & \multicolumn{2}{c}{D=30} \\\cline{2-5}
        & EBGWO & GWO   & EBGWO & GWO \\ \hline
    
    F1    & \textbf{2.57443E-01} & 5.33915E-01 & \textbf{6.78962E-01} & 9.19988E-01 \\
    F2    & \textbf{2.27636E-01} & 5.04466E-01 & \textbf{6.26778E-01} & 8.55688E-01 \\
    F3    & \textbf{2.26005E-01} & 5.07412E-01 & \textbf{5.99625E-01} & 7.80280E-01 \\
    F4    & \textbf{2.22712E-01} & 5.08896E-01 & \textbf{5.90939E-01} & 7.76646E-01 \\
    F5    & \textbf{2.45573E-01} & 5.39160E-01 & \textbf{6.50512E-01} & 8.54939E-01 \\
    F6    & \textbf{2.06016E+00} & 2.25300E+00 & \textbf{5.64472E+00} & 5.85205E+00 \\
    F7    & \textbf{2.71699E-01} & 5.22851E-01 & \textbf{6.76287E-01} & 9.10568E-01 \\
    F8    & \textbf{2.26198E-01} & 4.87149E-01 & \textbf{5.87757E-01} & 8.02861E-01 \\
    F9    & \textbf{2.37948E-01} & 4.91336E-01 & \textbf{6.22065E-01} & 8.56690E-01 \\
    F10   & \textbf{2.85948E-01} & 5.37783E-01 & \textbf{7.19027E-01} & 9.63514E-01 \\
    F11   & \textbf{2.99830E-01} & 5.73777E-01 & \textbf{8.07212E-01} & 1.05481E+00 \\
    F12   & \textbf{5.33946E-01} & 8.00030E-01 & \textbf{1.49308E+00} & 1.72033E+00 \\
    F13   & \textbf{2.28371E-01} & 4.67959E-01 & \textbf{5.96877E-01} & 8.14211E-01 \\
    F14   & \textbf{2.36427E-01} & 4.91210E-01 & \textbf{6.01025E-01} & 8.01521E-01 \\
    F15   & \textbf{2.44314E-01} & 5.04867E-01 & \textbf{6.48083E-01} & 8.04354E-01 \\
    F16   & \textbf{2.53027E-01} & 4.99860E-01 & \textbf{6.32409E-01} & 8.65807E-01 \\
    F17   & \textbf{2.71250E-01} & 5.50087E-01 & \textbf{7.59393E-01} & 9.93048E-01 \\
    F18   & \textbf{2.45329E-01} & 5.03560E-01 & \textbf{6.52141E-01} & 9.00193E-01 \\
    F19   & \textbf{6.05648E-01} & 8.59885E-01 & \textbf{1.86095E+00} & 2.32137E+00 \\
    F20   & \textbf{2.66903E-01} & 5.17061E-01 & \textbf{6.40480E-01} & 9.17434E-01 \\
    F21   & \textbf{2.61674E-01} & 5.38683E-01 & \textbf{6.76747E-01} & 1.04316E+00 \\
    F22   & \textbf{2.92598E-01} & 5.60898E-01 & \textbf{7.72988E-01} & 9.64225E-01 \\
    F23   & \textbf{5.06522E-01} & 7.69019E-01 & \textbf{1.45095E+00} & 1.66077E+00 \\
    F24   & \textbf{4.38957E-01} & 7.17328E-01 & \textbf{1.13358E+00} & 1.33951E+00 \\
    F25   & \textbf{4.69847E-01} & 7.37295E-01 & \textbf{1.33371E+00} & 1.54866E+00 \\
    F26   & \textbf{2.36236E+00} & 2.64601E+00 & 7.07515E+00 & \textbf{7.03532E+00} \\
    F27   & \textbf{2.41103E+00} & 2.72349E+00 & \textbf{6.75662E+00} & 7.02159E+00 \\
    F28   & \textbf{6.33840E-01} & 9.32547E-01 & \textbf{1.74896E+00} & 1.98052E+00 \\
    F29   & \textbf{8.08923E-01} & 1.12551E+00 & \textbf{2.24968E+00} & 2.49620E+00 \\
    F30   & \textbf{5.00398E-01} & 8.14709E-01 & \textbf{1.36442E+00} & 1.66345E+00 \\
    \bottomrule
    \end{tabular}}
\end{table}%

\begin{table}[htbp]
  \centering
  \caption{Computation time of EBGWO and GWO in 50D and 100D}
      \label{time2}
      \tabcolsep 6.5pt 
      \resizebox{3in}{43mm}{
    \begin{tabular}{c|cc|cc}
    \toprule
   \multicolumn{5}{c}{Computation Time} \\
    \hline
   \multirow{2}{*}{F}       &    \multicolumn{2}{c|}{D=50} & \multicolumn{2}{c}{D=100} \\\cline{2-5}
        & EBGWO & GWO   & EBGWO & GWO \\ \hline
    F1    & \textbf{1.10733E+00} & 1.25243E+00 & 2.64066E+00 & \textbf{2.55704E+00} \\
    F2    & \textbf{9.90340E-01} & 1.21593E+00 & \textbf{2.28161E+00} & 2.33669E+00 \\
    F3    & \textbf{1.00878E+00} & 1.18842E+00 & 2.45394E+00 & \textbf{2.31238E+00} \\
    F4    & \textbf{1.02678E+00} & 1.19147E+00 & 2.45920E+00 & \textbf{2.31397E+00} \\
    F5    & \textbf{1.11429E+00} & 1.27215E+00 & 2.61002E+00 & \textbf{2.50889E+00} \\
    F6    & 9.78085E+00 & \textbf{9.66094E+00} & 1.97146E+01 & \textbf{1.96098E+01} \\
    F7    & \textbf{1.16203E+00} & 1.30733E+00 & 2.76805E+00 & \textbf{2.62280E+00} \\
    F8    & \textbf{8.65649E-01} & 1.10380E+00 & 2.00373E+00 & \textbf{1.86970E+00} \\
    F9    & \textbf{1.09525E+00} & 1.24415E+00 & 2.60504E+00 & \textbf{2.49025E+00} \\
    F10   & \textbf{1.18903E+00} & 1.36804E+00 & 2.47828E+00 & \textbf{2.36112E+00} \\
    F11   & \textbf{1.34298E+00} & 1.51597E+00 & \textbf{2.71581E+00} & 2.85564E+00 \\
    F12   & \textbf{2.63889E+00} & 2.74563E+00 & 5.57998E+00 & \textbf{5.43131E+00} \\
    F13   & \textbf{9.60947E-01} & 1.11980E+00 & 2.44978E+00 & \textbf{2.32971E+00} \\
    F14   & \textbf{9.80441E-01} & 1.16004E+00 & 2.47794E+00 & \textbf{2.34972E+00} \\
    F15   & \textbf{1.16255E+00} & 1.23821E+00 & 2.66400E+00 & \textbf{2.50154E+00} \\
    F16   & \textbf{1.13035E+00} & 1.28662E+00 & 2.69644E+00 & \textbf{2.57132E+00} \\
    F17   & \textbf{1.13896E+00} & 1.32810E+00 & \textbf{2.79136E+00} & 2.82873E+00 \\
    F18   & \textbf{1.08577E+00} & 1.19082E+00 & \textbf{2.53047E+00} & 2.55790E+00 \\
    F19   & \textbf{2.87653E+00} & 3.17533E+00 & \textbf{6.10037E+00} & 6.17802E+00 \\
    F20   & \textbf{1.08752E+00} & 1.24580E+00 & \textbf{2.49016E+00} & 2.64948E+00 \\
    F21   & \textbf{1.13191E+00} & 1.31664E+00 & \textbf{2.59050E+00} & 2.67555E+00 \\
    F22   & \textbf{1.32649E+00} & 1.53390E+00 & \textbf{2.95511E+00} & 2.97198E+00 \\
    F23   & \textbf{2.63510E+00} & 2.65947E+00 & \textbf{6.23270E+00} & 6.51885E+00 \\
    F24   & \textbf{2.04942E+00} & 2.16884E+00 & \textbf{6.29477E+00} & 6.36404E+00 \\
    F25   & \textbf{2.30654E+00} & 2.47679E+00 & \textbf{5.77246E+00} & 5.90779E+00 \\
    F26   & \textbf{1.17171E+01} & 1.18476E+01 & \textbf{2.55205E+01} & 2.58586E+01 \\
    F27   & 1.22375E+01 & \textbf{1.22098E+01} & \textbf{2.52405E+01} & 2.54313E+01 \\
    F28   & \textbf{3.34630E+00} & 3.60052E+00 & \textbf{8.18350E+00} & 8.23793E+00 \\
    F29   & \textbf{4.12876E+00} & 4.38850E+00 & \textbf{8.91686E+00} & 8.99851E+00 \\
    F30   & \textbf{2.38418E+00} & 2.59502E+00 & 5.62204E+00 & \textbf{5.60089E+00} \\
    \bottomrule
    \end{tabular}}
\end{table}%

\section{Experiments and discussion}\label{sec:experiments}
Three experiments are conducted in this section to evaluate the EBGWO algorithm. In Experiment 1, the convergence effect of EBGWO is tested on the benchmark function suite, and an ablation study is provided to verify the performance of the core mechanisms. In Experiment 2, a statistical analysis is designed to verify the hypothesis of this research. In Experiment 3, three engineering examples are used to verify the algorithm's feasibility in addressing practical problems.

\subsection{Experiment setting}
\textbf{Data sets.} The classic IEEE CEC 2014 benchmark functions \citep{27} are used to evaluate the performance of the proposed algorithm in this section. The dataset listed in Table \ref{CEC} incorporates 30 test benchmark functions, including unimodal functions, simple multimodal functions, hybrid functions, and composition functions. Functions F1 to F3 are unimodal, F4 to F16 are simple multimodal, F17 to F22 are hybrid, and F23 to F30 are composite. This suite of benchmark functions is widely used to evaluate the performance of different meta-heuristics \citep{18,23,28,29}. For each function type, the optimizer performs 30 runs in four different dimensions (10Dim, 30Dim, 50Dim, and 100Dim) as per IEEE CEC 2014 guidelines.

\begin{table}[htbp]
\begin{center}
\caption{\label{CEC} Benchmark functions of IEEE CEC 2014}
\resizebox{4.8in}{90mm}{
\begin{tabular}{cc}
\hline
\textbf{A. Unimodal Functions:}\\
\hline
Rotated High Conditioned Elliptic Function & $F_1(x)=f_1(M(x-o_1))+100$\\
Rotated Bent Cigar Function & $F_2(x)=f_2(M(x-o_2))+200$\\
Rotated Discus Function & $F_3(x)=f_3(M(x-o_3))+300$\\
\hline
\textbf{B. Multimodal Functions:}\\
\hline
Shifted and Rotated Rosenbrock’s Function & $F_4(x)=f_4(M(\dfrac{2.048(x-o_4)}{100})+1)+400$\\
Shifted and Rotated Ackley’s Function & $F_5(x)=f_5(M(x-o_5))+500$\\
Shifted and Rotated Weierstrass Function & $F_6(x)=f_6(M(\dfrac{0.5(x-o_6)}{100}))+600$\\
Shifted and Rotated Griewank’s Function & $F_7(x)=f_7(M(\dfrac{600(x-o_7)}{100}))+700$\\
Shifted Rastrigin’s Function & $F_8(x)=f_8(M(\dfrac{5.12(x-o_8)}{100}))+800$\\
Shifted and Rotated Rastrigin’s Function & $F_9(x)=f_8(M(\dfrac{5.12(x-o_9)}{100}))+900$\\
Shifted Schwefel’s Function & $F_{10}(x)=f_9(M(\dfrac{1000(x-o_{10})}{100}))+1000$\\
Shifted and Rotated Schwefel’s Function & $F_{11}(x)=f_9(M(\dfrac{1000(x-o_{11})}{100}))+1100$\\
Shifted and Rotated Katsuura Function & $F_{12}(x)=f_{10}(M(\dfrac{5(x-o_{12})}{100}))+1200$\\
Shifted and Rotated HappyCat Function & $F_{13}(x)=f_{11}(M(\dfrac{5(x-o_{13})}{100}))+1300$\\
Shifted and Rotated HGBat Function & $F_{14}(x)=f_{12}(M(\dfrac{5(x-o_{14})}{100}))+1400$\\
Shifted and Rotated Expanded Griewank’s plus Rosenbrock’s Function & $F_{15}(x)=f_{13}(M(\dfrac{5(x-o_{15})}{100})+1)+1500$\\
Shifted and Rotated Expanded Scaffer’s F6 Function & $F_{16}(x)=f_{14}(M((x-o_{16}))1)+1600$\\
\hline
\textbf{Hybrid Functions}\\
\hline
$F_{17}=f_9(M_1Z_1)+f_8(M_2Z_2)+f_3(M_3Z_3)+1700$ & p=[0.3,0.3,0.4]\\
$F_{18}=f_2(M_1Z_1)+f_8(M_2Z_2)+f_3(M_3Z_3)+1800$ & p=[0.3,0.3,0.4]\\
$F_{19}=f_7(M_1Z_1)+f_8(M_2Z_2)+f_3(M_3Z_3)+f_8(M_4Z_4)+1900$ & p=[0.2,0.2,0.3,0.3]\\
$F_{20}=f_{12}(M_1Z_1)+f_3(M_2Z_2)+f_{13}(M_3Z_3)+f_8(M_4Z_4)+2000$ & p=[0.2,0.2,0.3,0.3]\\
$F_{21} = f_{14}(M_1Z_1)+f_{12}(M_2Z_2)+f_4(M_3Z_3) +f_9(M_4Z_4)+f_1(M_5Z_5)+2100$& p =[0.1,0.2,0.2,0.2, 0.3]\\
$F_{22} = f_{10}(M_1Z_1)+f_{11}(M_2Z_2)+f_{13}(M_3Z_3) +f_9(M_4Z_4)+f_5(M_5Z_5)+2200$& p =[0.1,0.2,0.2,0.2, 0.3]\\
\textbf{Notes:}\\
$Z_1=[y_{s_1},y_{s_1}......,y_{s_{n1}}]$\\
$Z_2=[y_{s_{n1+1}},y_{s_{n1+2}}......,y_{s_{n1+n2}}]$\\
$Z_N=[y_{s_{\sum_{i=1}^{N-1} n1+1}},y_{s_{\sum_{i=1}^{N-1} n1+2}}......,y_{s_{5D}}]$\\
$y=x-o_i$,$S=randperm(1:D)$,$percentageofg_{i}(x)$\\
$n_1=[p_1D]$,$n_2=[p_2D]$,...,$n_{N-1}=[p_{N-1}D]$,$n_N=D-\sum_{i=1}^ {N-1}n+i$\\
\hline
\textbf{Composition Functions}\\
\hline
$F_{23}=w_1*F'_4(x)+w_2*[1e^{-6}F'_1(x)+100]+w_3*[1e^{-26}F'_2(x)+200]$\\$+w_4*[1e^{-6}F'_3(x)+300]+w_5*[1e^{-6}F'_1(x)+400]+2300$ &  $\sigma$ =[10,20,30,40,50]\\
$F_{24}=w_1*F'_{10}(x)+w_2*[F'_9(x)+100]+w_3*[F'_{14}(x)+200]+2400$&  $\sigma$ =[20,20,20]\\
$F_{25}=w_1*0.25F'_{11}(x)+w_2*[F'_9(x)+100]+w_3*[1e^{-7}F'_1(x)+200]+2500$&  $\sigma$ =[10,30,50]\\
$F_{26}=w_1*0.25F'_{11}(x)+w_2*[F'_{13}(x)+100]+w_3*[1e^{-7}F'_1(x)+200]$\\$+w_4*[2.5F'_6(x)+300]+w_5*[1e^{-6}F'_{13}(x)+400]+2700$ &  $\sigma$ =[10,10,10,10,10]\\
$F_{27}=w_1*10F'_{14}(x)+w_2*[10F'_{9}(x)+100]+w_3*[2.5F'_{1}1(x)+200]$\\$+w_4*[25F'_{16}(x)+300]+w_5*[1e^{-6}F'_{1}(x)+400]+2700$ &  $\sigma$ = =[10,10,10,20,20]\\
$F_{28}=w_1*2.5F'_{15}(x)+w_2*[10F'_{9}(x)+100]+w_3*[2.5F'_{1}1(x)+200]$\\$+w_4*[5e^{-4}F'_{16}(x)+300]+w_5*[1e^{-6}F'_{1}(x)+400]+2800$ &  $\sigma$ = =[10,20,30,40,50]\\
$F_{29}=w_1*F'_{17}(x)+w_2*[F'_{18}(x)+100]+w_3*[F'_{19}(x)+200]+2900$&  $\sigma$ =[10,30,50]\\
$F_{30}=w_1*F'_{20}(x)+w_2*[F'_{21}(x)+100]+w_3*[F'_{22}(x)+200]+3000$&  $\sigma$ =[10,30,50]\\
\textbf{Notes:}\\
$w_i=\dfrac{1}{\sqrt{\sum_{j=1}^D(x_j-o_{ij})}}exp(-\dfrac{\sum_{j=1}^D(x_i-o_{ij}^2)}{2D \sigma_i^2})$\\
\hline
\end{tabular}}
\end{center}
\end{table}

\textbf{Environment.} The environment utilizes a computer equipped with an AMD A8-7100 Radeon R5, 8 Compute Cores 4C+4G, 1.08 GHz, and 4GB RAM. The codes are implemented using Matlab R2018a on a Windows 8 operating system.

\textbf{Evaluation Metrics.} The results of Experiment 1 are quantified using the mean value and standard deviation. In addition, the paper uses search history, trajectory history, average fitness, and convergence curve figures for qualitative analysis. For qualitative analysis and to determine the superior algorithm, we employ a ranking system based on the (w/t/l) values. The (w/t/l) values refer to the number of optimal solutions (wins), identical solutions (ties), and suboptimal solutions (losses) derived from testing each algorithm compared with other baselines. Additionally, we introduce an indicator, \textbf{Overall Effectiveness}, to evaluate the EBGWO algorithm's performance \citep{30}.

The equation of overall effectiveness is given in Eq. (\ref{23}).

\begin{equation}
\label{23}
\text{Overall Effectiveness} =\frac{(N-L)}{N} \times 100 \%
\end{equation}

where $N$ is the total number of Experiment 1 (120), and $L$ is the lose number of each compared algorithm.

\textbf{Baselines.} The proposed EBGWO algorithm is compared against six algorithms to evaluate the performance. These baseline algorithms incorporate original GWO \citep{5}, SOGWO \citep{13}, mGWO \citep{14}, AGWO \citep{18}, SCA \citep{31} and WOA \citep{32} algorithms. 

\begin{itemize}
\item \textbf{GWO \citep{5}:} Mirjalili et al. introduce the swarm intelligence algorithm inspired by the behaviour of grey wolves in nature. The GWO algorithm is recongnized for its excellent exploitation performance.
\item \textbf{SOGWO \citep{13}:} Dhargupta et al. combine the opposition-based learning method on weak wolves in selective dimensions with the GWO algorithm to improve the exploration capability of the algorithm. The SOGWO algorithm, designed in 2020, has been cited over 70 times in the Science Direct database for comparison. 
\item \textbf{mGWO \citep{14}:} Mittal et al. adopt an exponential decay function to enhance the exploration process in GWO, effectively improving its performance. The function is given in Eq. (\ref{24}). 
\begin{equation}
\label{24} 
a=2(1-\dfrac{t^{2}}{T^{2}}).
\end{equation}
\item \textbf{AGWO \citep{18}:} Meng et al. propose an advanced grey wolf optimization algorithm with elastic, circling, and attacking mechanisms in 2021, which aims to solve the local optimum and premature convergence problems in multilayer perceptron application. 
\item \textbf{SCA \citep{31}:} Mirjalili proposes the population-based Sine Cosine Algorithm to solve optimization problems. Due to its superior performance over PSO \citep{PSO}, GSA \citep{GSA}, and other renowned algorithms on the benchmark function suite, SCA is used as a comparison algorithm. It was cited by more than 1,800 papers in the Web of Science database. 
\item \textbf{WOA \citep{32}:} Mirjalili introduces the Whale Optimization Algorithm and applies it to six constrained engineering design problems. It exhibits strong performance in benchmark function tests, outperforming well-known algorithms such as PSO \citep{PSO} and DE \citep{DE}. It was cited by over 4000 papers in the Web of Science database. Therefore, it is a competitive swarm intelligence algorithm.
\end{itemize}

\textbf{Experiment Parameters.} To ensure a fair comparison, the population size for all algorithms in the experiments is set to 30, with a maximum iteration number of 500. The range of search boundaries for each variable is [-100, 100]. The termination criterion is the maximum number of iterations ($MaxIterations$). Table \ref{parameter} provides the details of the parameter setting for algorithms.

\begin{table}[htbp]
\begin{center}
  \caption{\label{parameter}Parameter setting for algorithms}
         \resizebox{3in}{25mm}{
    \begin{tabular}{ccl}
    \toprule
     Algorithms & parameters & values\\
    \hline
    \multirow{2}[1]{*}{GWO} & $a$ & linearly decreased from 2 to 0\\
      & $r_{1}, r_{2}$ & randomly selected in [0,1] \\
          \multirow{2}[1]{*}{EBGWO} & $a$ & linearly decreased from 2 to 0 \\
      & $r_{1}, r_{2}$ & randomly selected in [0,1]\\
     \multirow{2}[0]{*}{mGWO} & $a$ & $a=2(1-\frac{t^2}{T^2})$ \\
      & $r_{1}, r_{2}$ & randomly selected in [0,1] \\
    \multirow{2}[0]{*}{SOGWO} & $a$ & linearly decreased from 2 to 0 \\
      & $r_{1}, r_{2}$ & randomly selected in [0,1] \\
    \multirow{2}[0]{*}{AGWO} & $a$ & linearly decreased from 2 to 0 \\
      & $r_{1}, r_{2}$ & randomly selected in [0,1] \\
\multirow{2}[0]{*}{SCA} & $a$ & 2 \\
      & $r_{1}$ & linearly decreased from a to 0 \\
    \multirow{4}[0]{*}{WOA} & $a$ & linearly decreased from 2 to 0 \\
      & $p$ & randomly selected in [0,1] \\
      & $l$ & randomly selected in [-1,1] \\
      & $r$ & randomly selected in [0,1] \\
    \bottomrule
    \end{tabular}}%
    \end{center}
\end{table}%

\subsection{Experiment 1: Benchmark Functions Suite Analysis}
The aim of Experiment 1 is to demonstrate the convergence of the EBGWO algorithm by comparing its performance with the baseline algorithms. In this experiment, we calculate the mean, standard deviation, overall effectiveness and ranking of the convergence results for each algorithm. The simulation experiment results are explained as follows. 

\subsubsection{Qualitative Analysis}
\begin{figure}[ht]
\centering
\subfigure{
\includegraphics[width=2in]{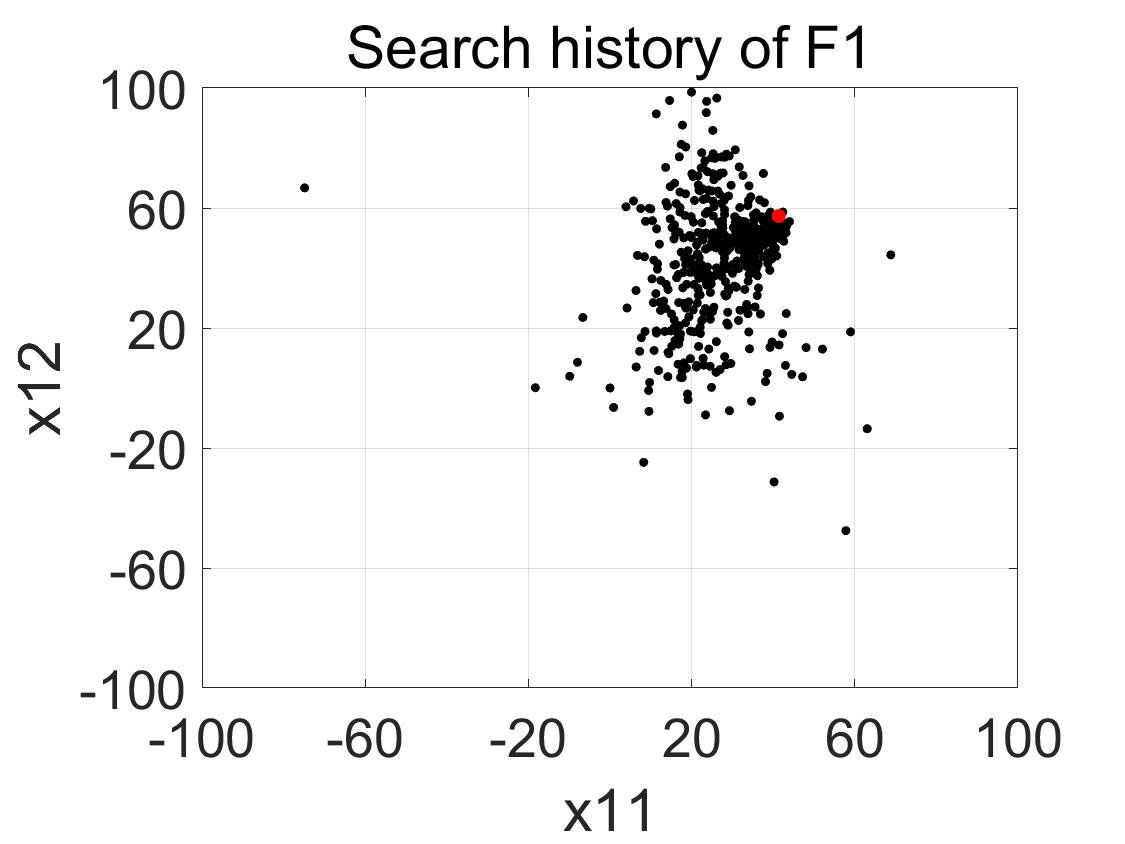}   
\label{gj:1-1}}
\subfigure{
\includegraphics[width=2in]{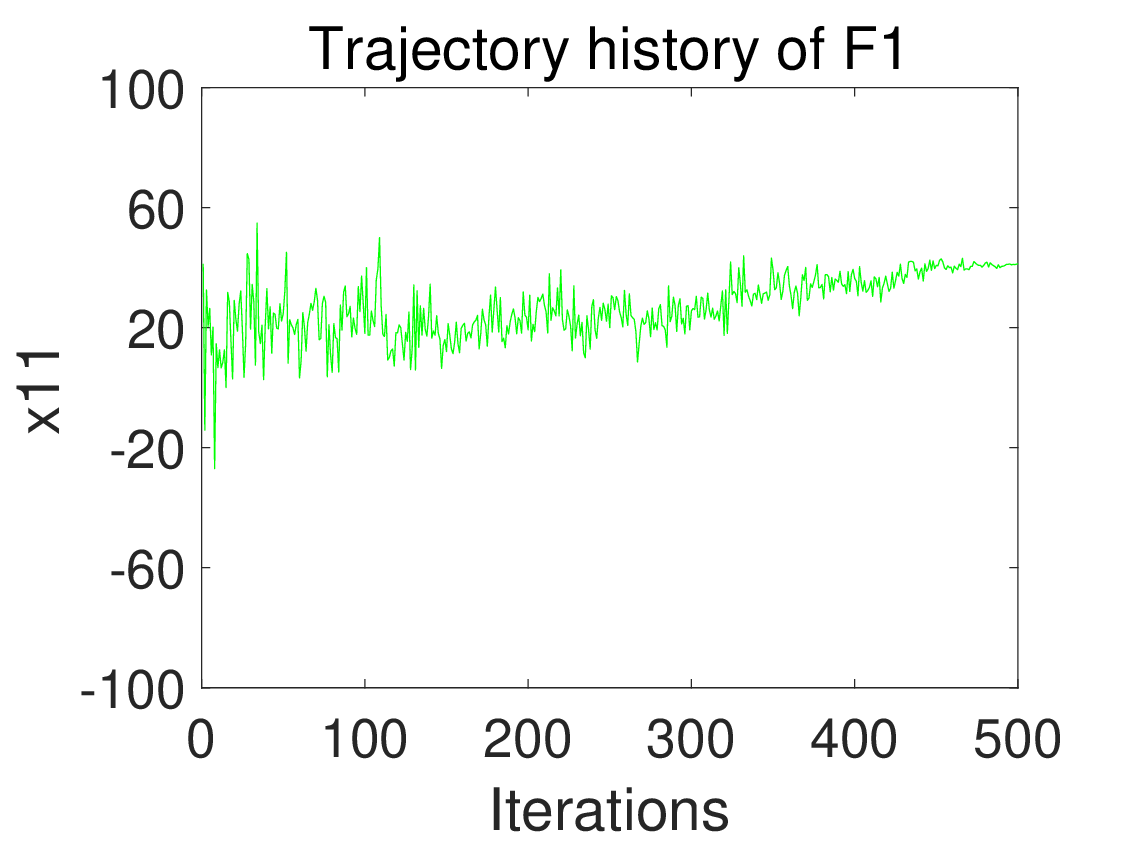}    
\label{gj:1-2}}
\subfigure{
\includegraphics[width=2in]{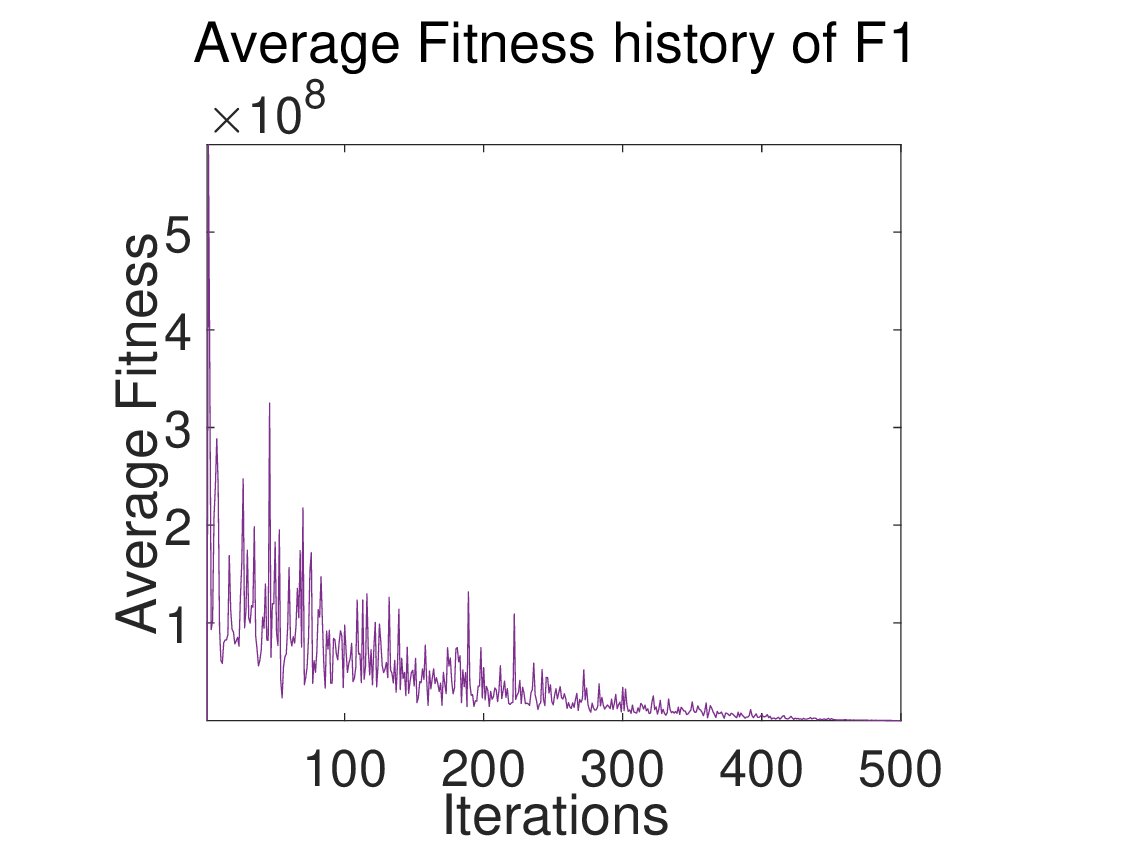}    
\label{gj:1-3}}
\subfigure{
\includegraphics[width=2in]{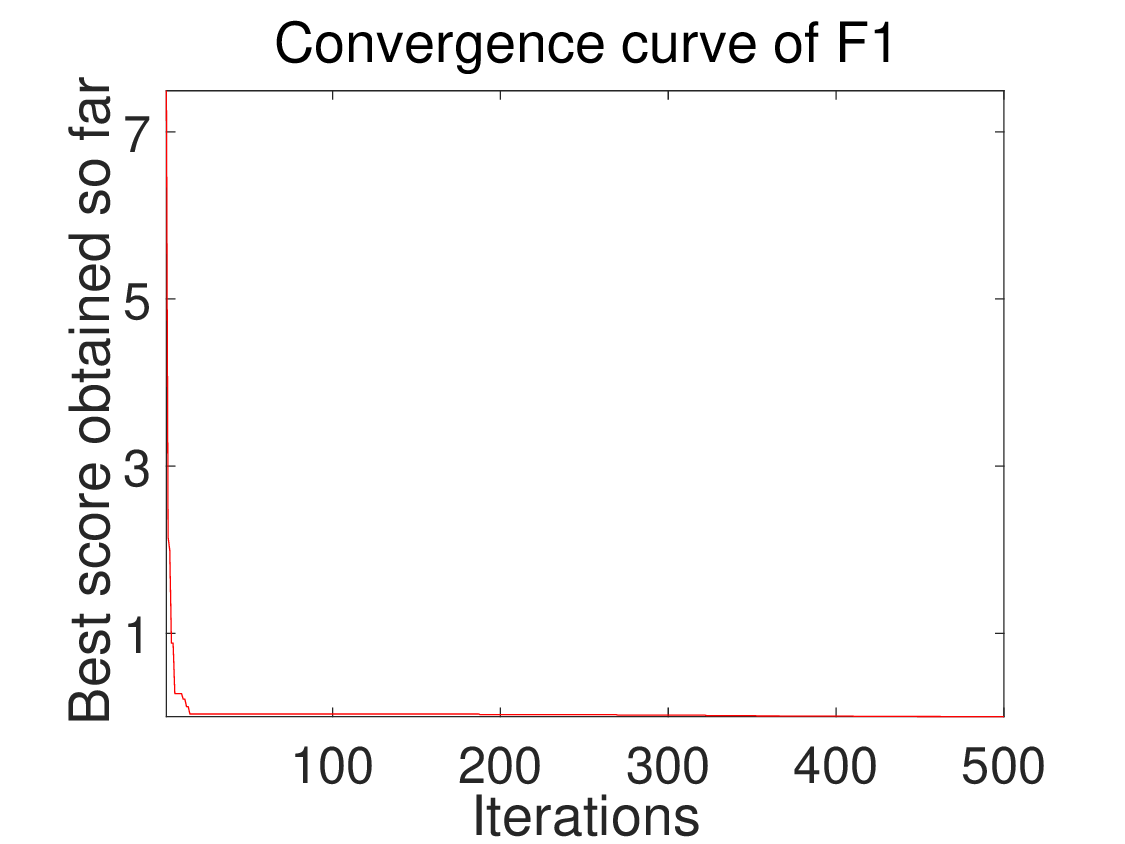}    
\label{gj:1-4}}

\caption{Search history (a), trajectory history (b), average fitness (c) and convergence curve (d) of EBGWO of $F1$}
\label{gj}
\end{figure}

We use search history, trajectory history, average fitness, and convergence curve figures to evaluate the performance of EBGWO. Fig. \ref{gj} presents the search results of EBGWO on Function 1. Fig. \ref{gj:1-1} demonstrates the final distribution of the population after several iterations. The search wolves get close to the prey, encircle and attack the prey. Fig. \ref{gj:1-2} depicts the trajectory history of the population, and the search wolves explore the global search space. Then they exploit the local search region to find the prey. The average fitness history and convergence curve indicate that the EBGWO algorithm exhibits strong convergence capability and can obtain high-quality solutions.

Furthermore, Figs. \ref{cfig:10} to \ref{cfig:100} present the convergence curve figures used to compare the EBGWO algorithm with other algorithms. The proposed EBGWO algorithm is compared with six other algorithms, including GWO, mGWO, SOGWO, AGWO, SCA and WOA algorithms. In these figures, the convergence curves are plotted by the mean values of the best solution in each iteration over 30 runs \citep{28}. As evident from these figures, the EBGWO algorithm demonstrates faster convergence speed and robust searching capabilities across various types of functions.

\subsubsection{Quantitative analysis}
\begin{figure}[ht]
\centering
\subfigure[$f_2$ ]{
\includegraphics[width=2in]{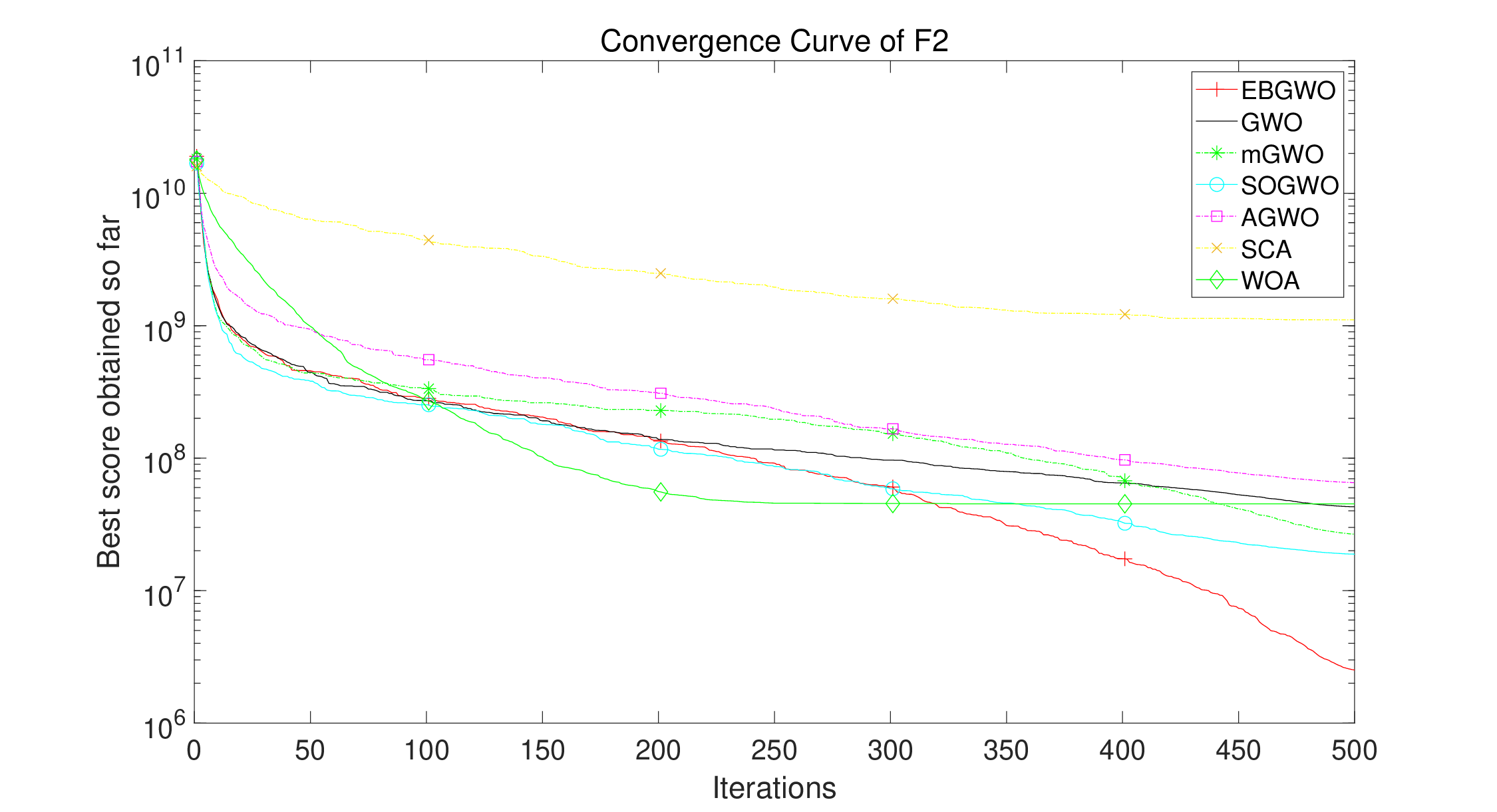}   
\label{cfig:10-1}}
\subfigure[$f_{11}$  ]{
\includegraphics[width=2in]{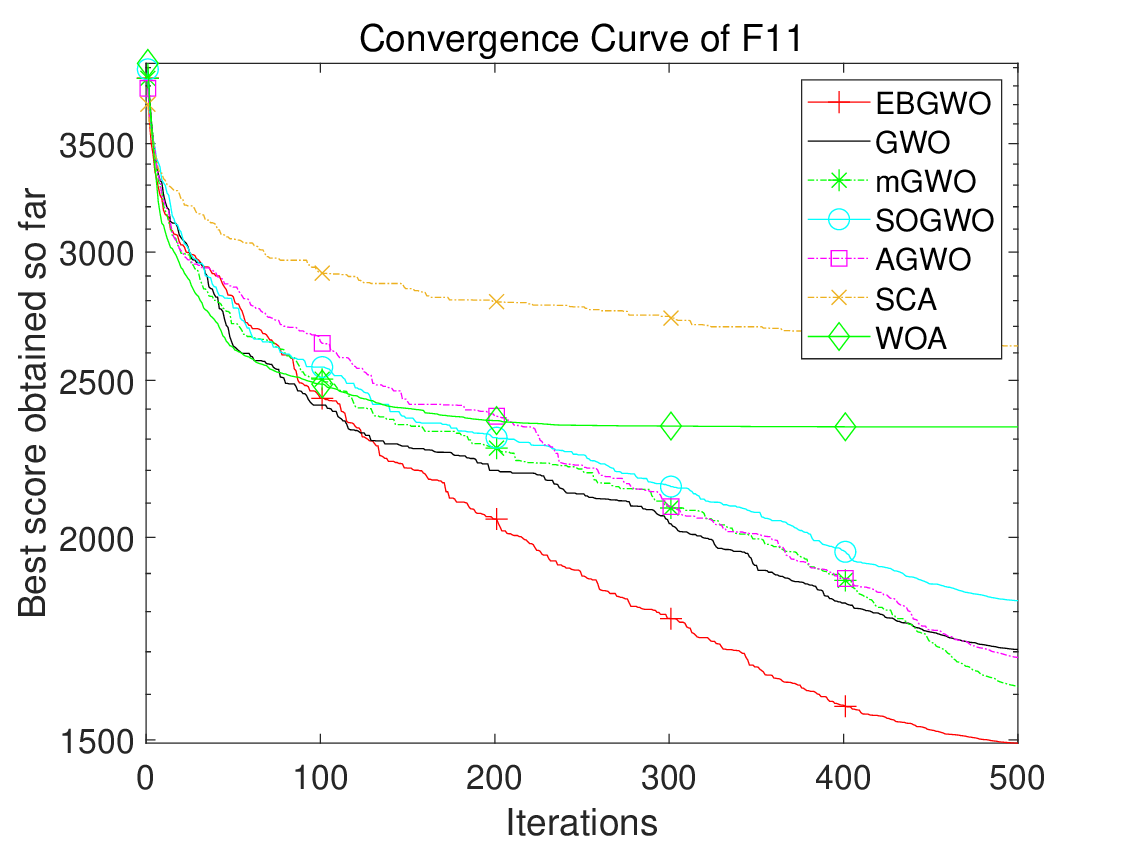}    
\label{cfig:10-2}}
\subfigure[$f_{17}$  ]{
\includegraphics[width=2in]{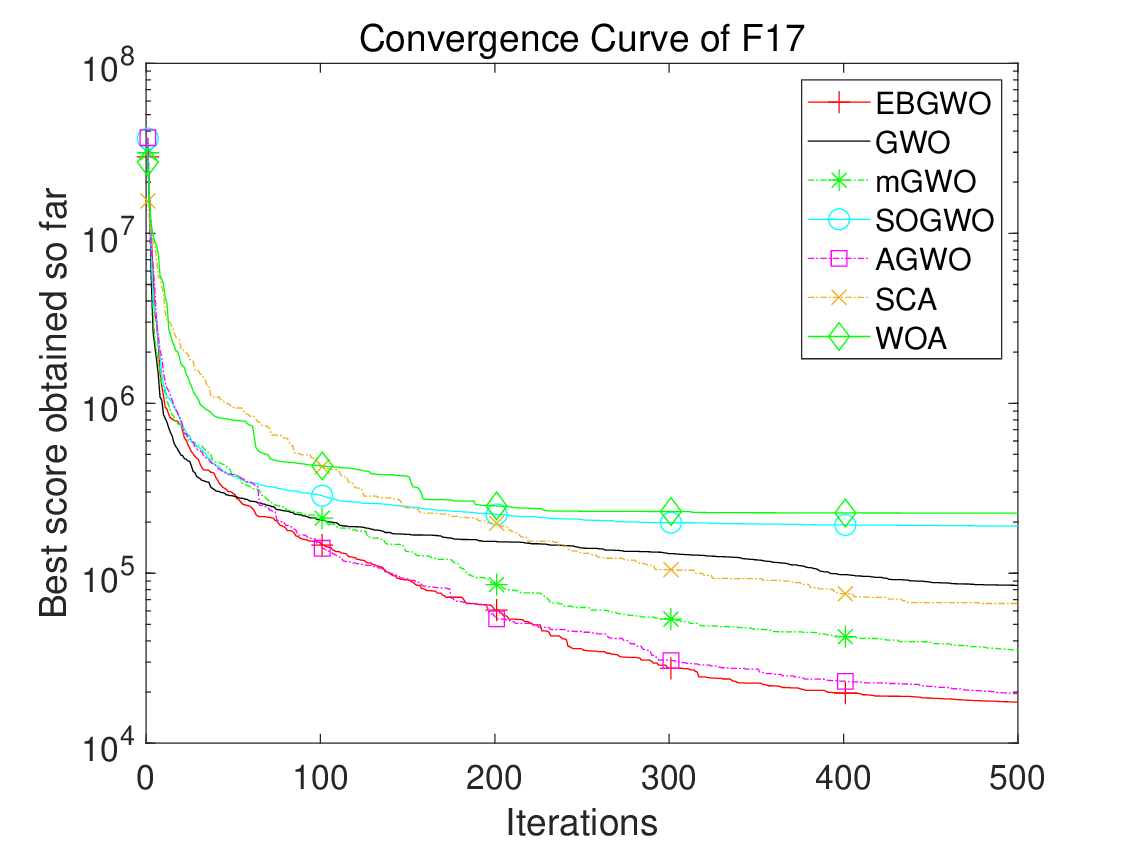}    
\label{cfig:10-3}}
\subfigure[$f_{28}$  ]{
\includegraphics[width=2in]{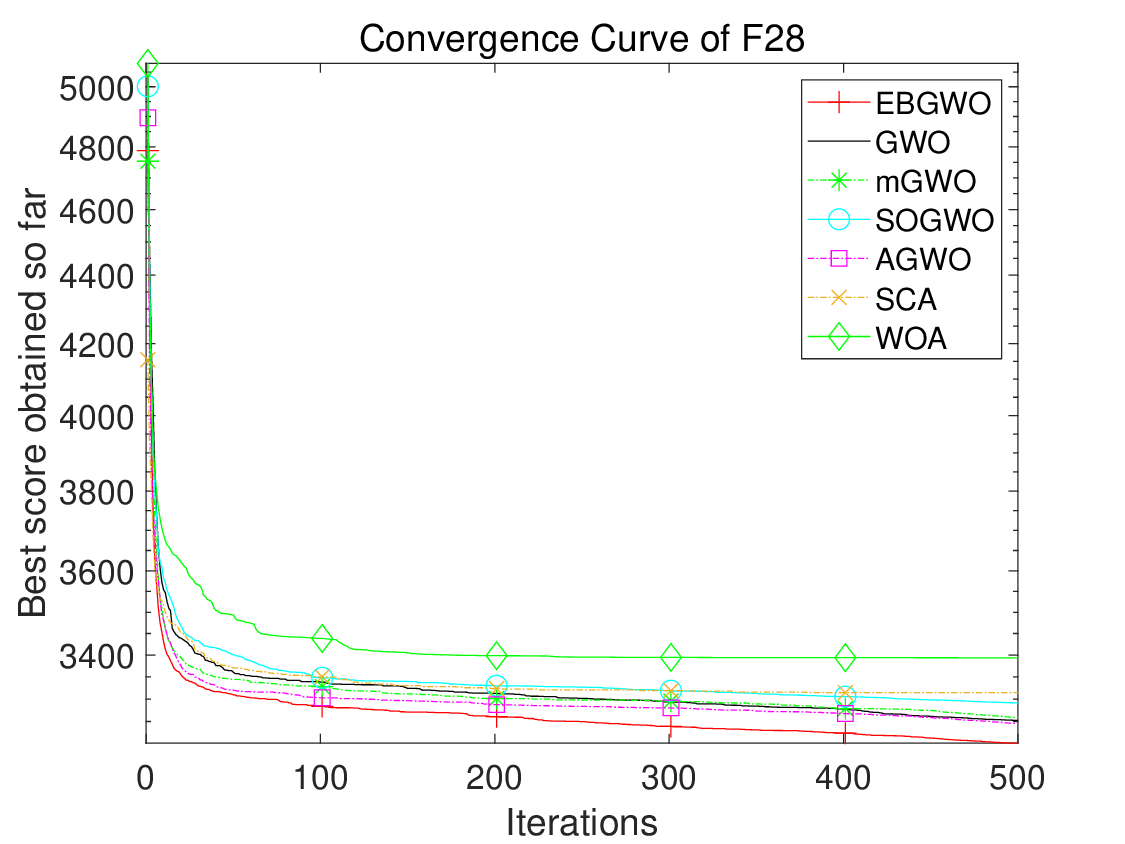}    
\label{cfig:10-4}}
\hspace{0.02\textwidth}
\caption{Convergence curves of EBGWO and other algorithms, Dim=10}
\label{cfig:10}
\end{figure}

\begin{figure}[ht]
\centering
\subfigure[$f_2$  ]{
\includegraphics[width=2in]{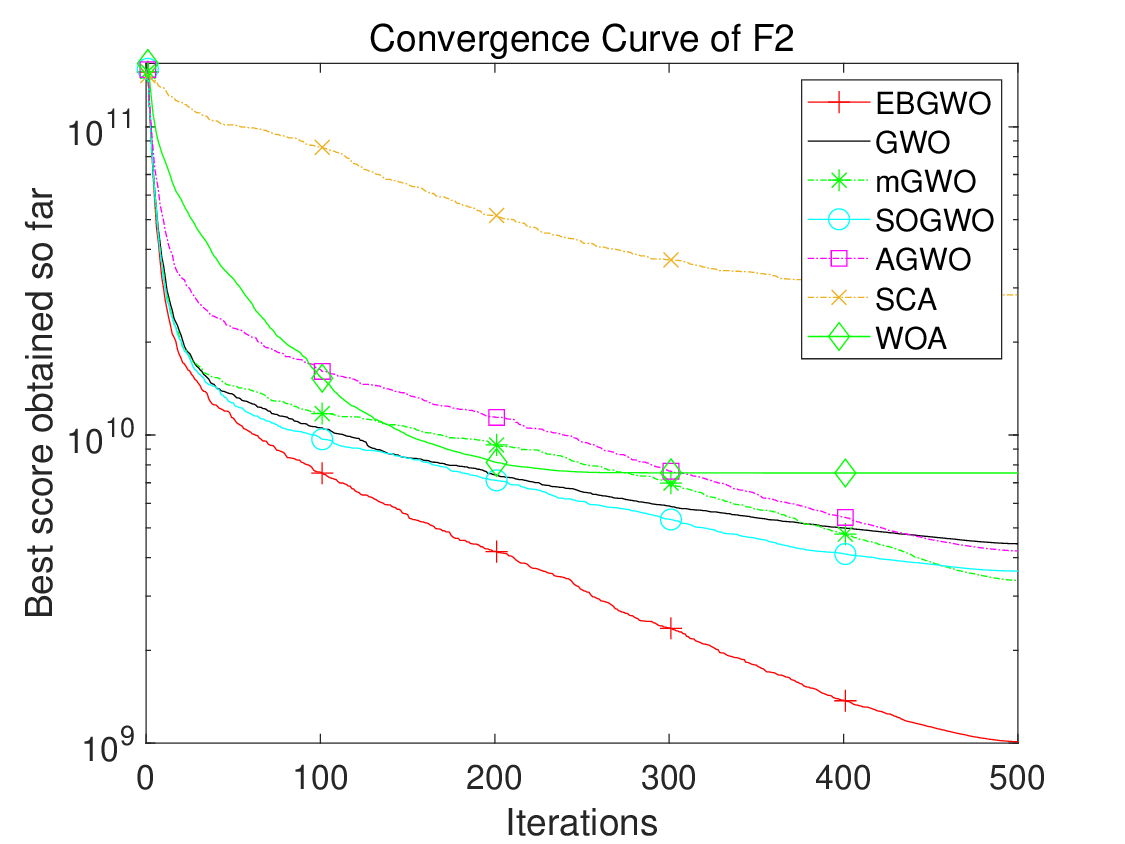}   
\label{cfig:30-1}}
\subfigure[$f_{11}$ ]{
\includegraphics[width=2in]{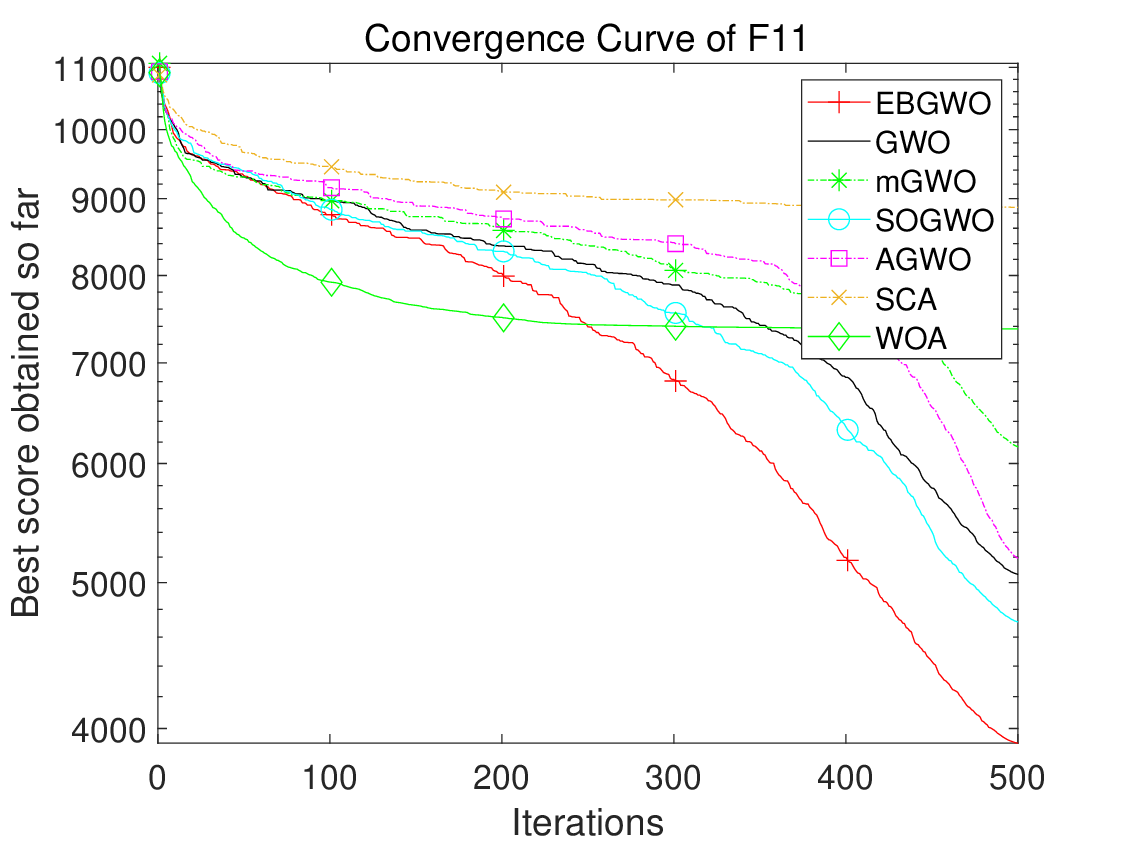}    
\label{cfig:30-2}}
\subfigure[$f_{17}$  ]{
\includegraphics[width=2in]{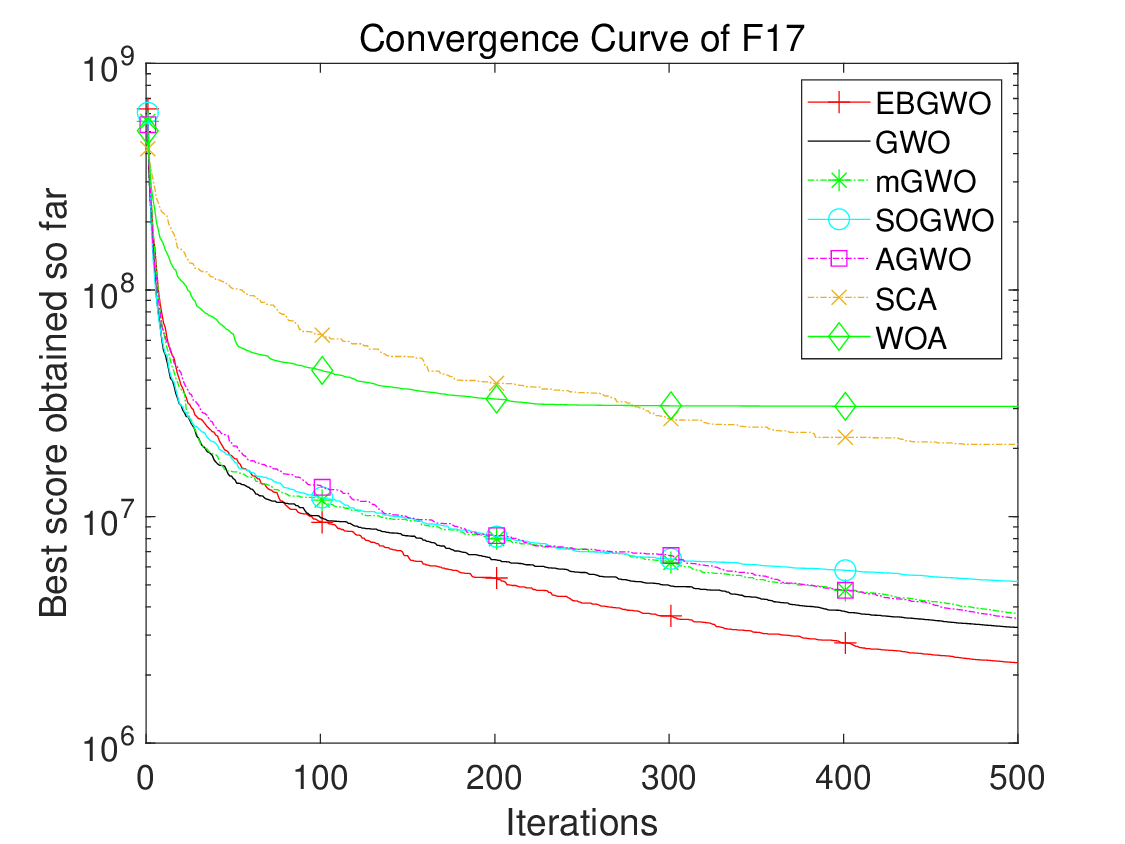}    
\label{cfig:30-3}}
\subfigure[$f_{30}$]{
\includegraphics[width=2in]{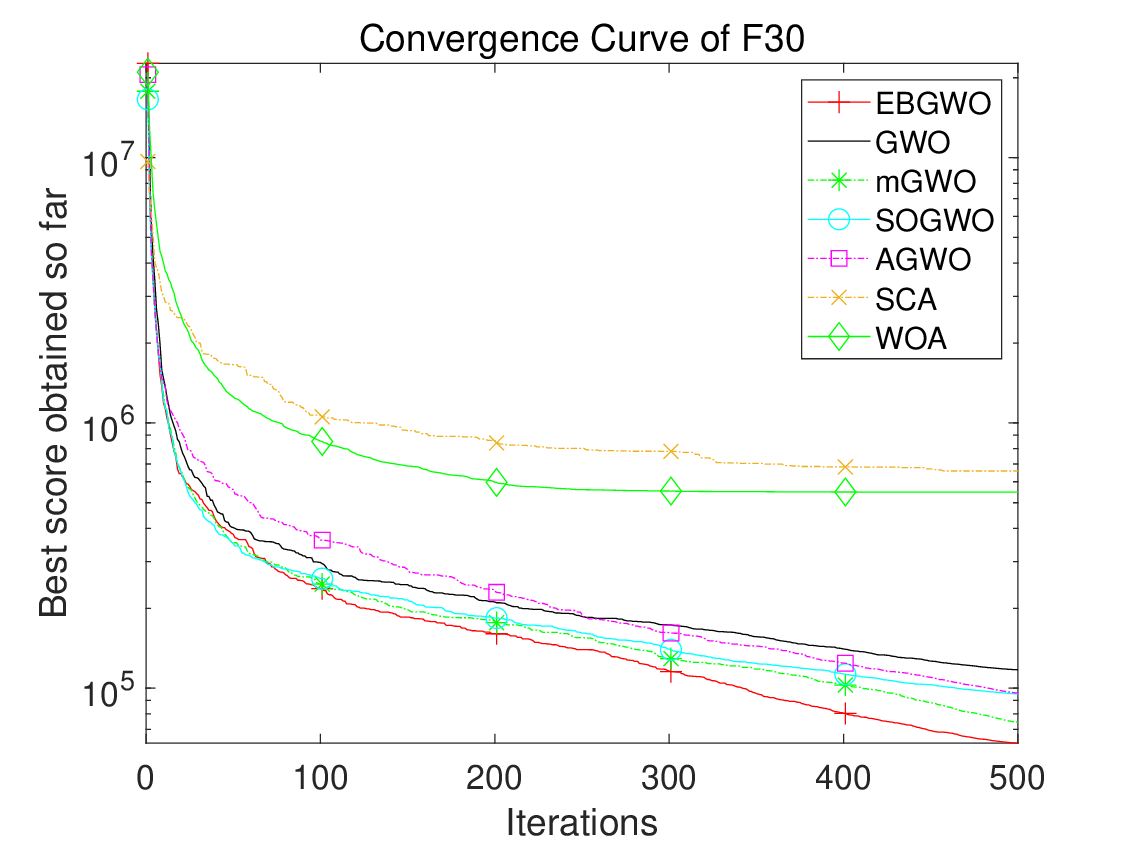}    
\label{cfig:30-4}}
\hspace{0.02\textwidth}
\caption{Convergence curves of EBGWO and other algorithms, Dim=30}
\label{cfig:30}
\end{figure}

\begin{figure}[ht]
\centering
\subfigure[$f_1$  ]{
\includegraphics[width=2in]{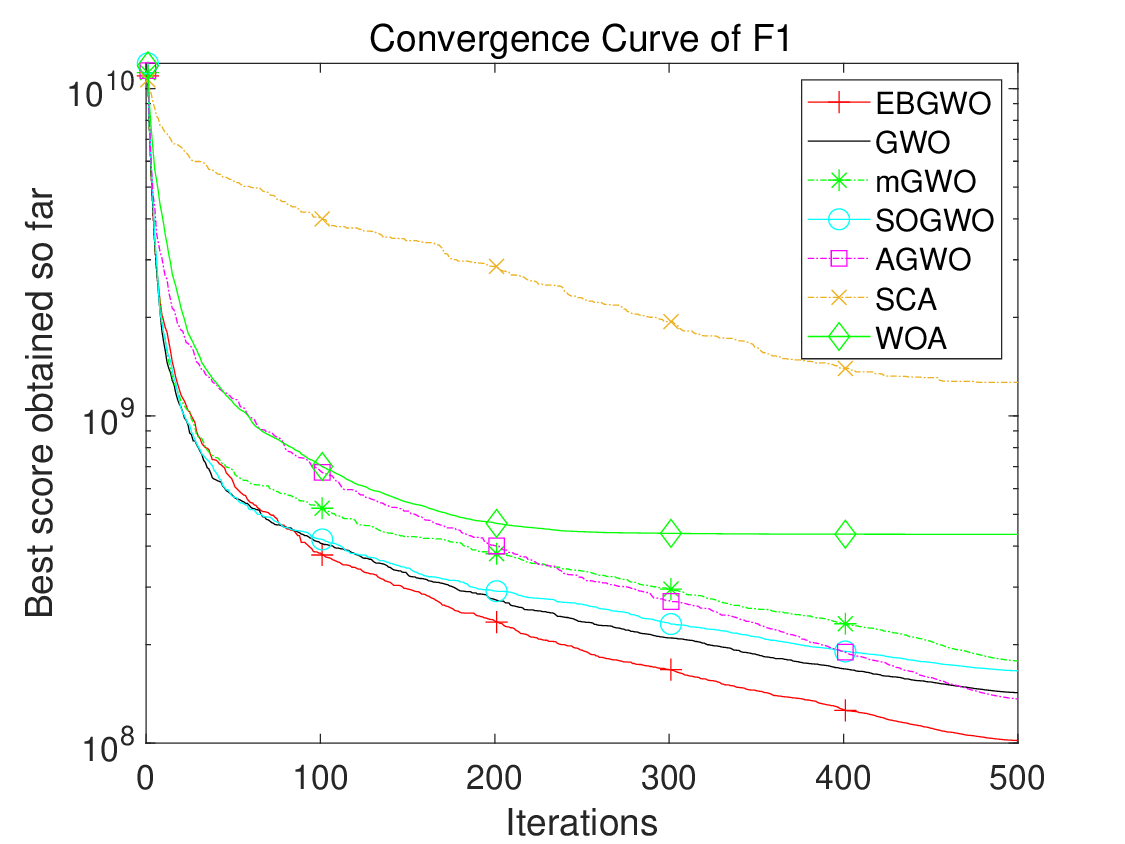}   
\label{cfig:50-1}}
\subfigure[$f_{13}$  ]{
\includegraphics[width=2in]{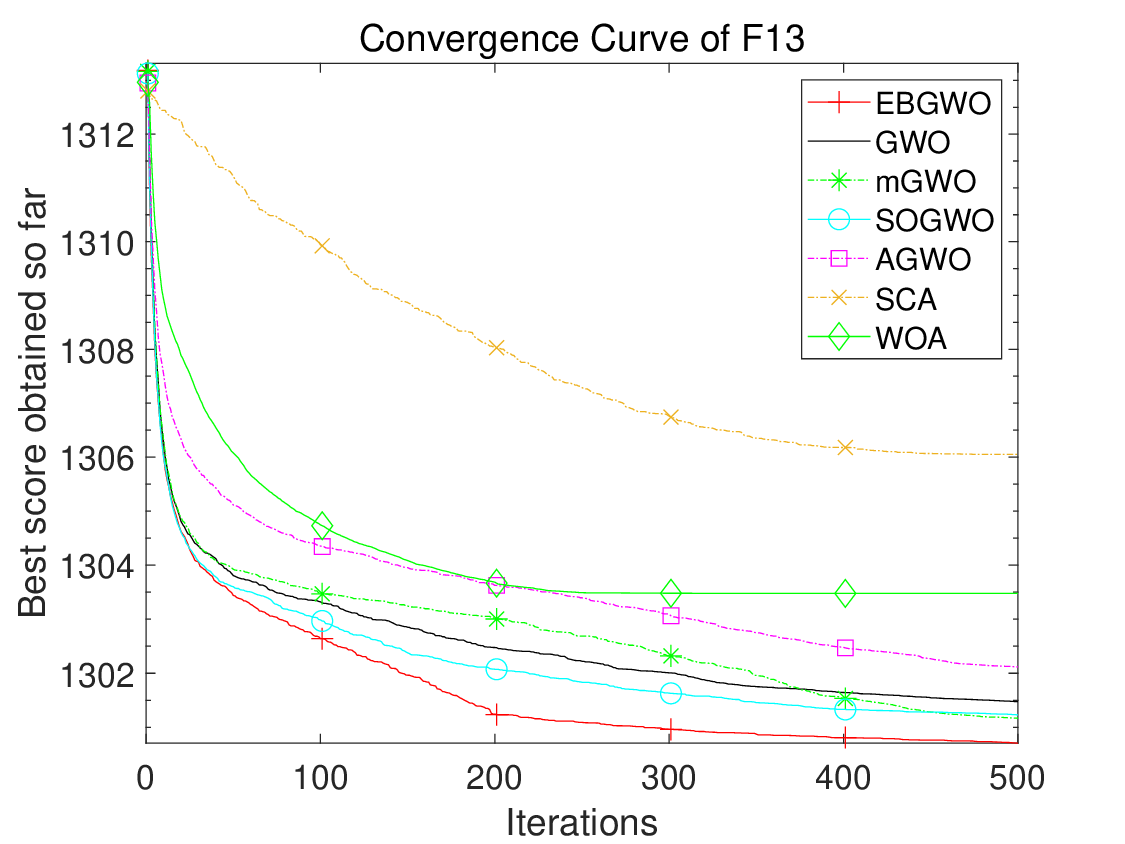}    
\label{cfig:50-2}}
\subfigure[$f_{17}$  ]{
\includegraphics[width=2in]{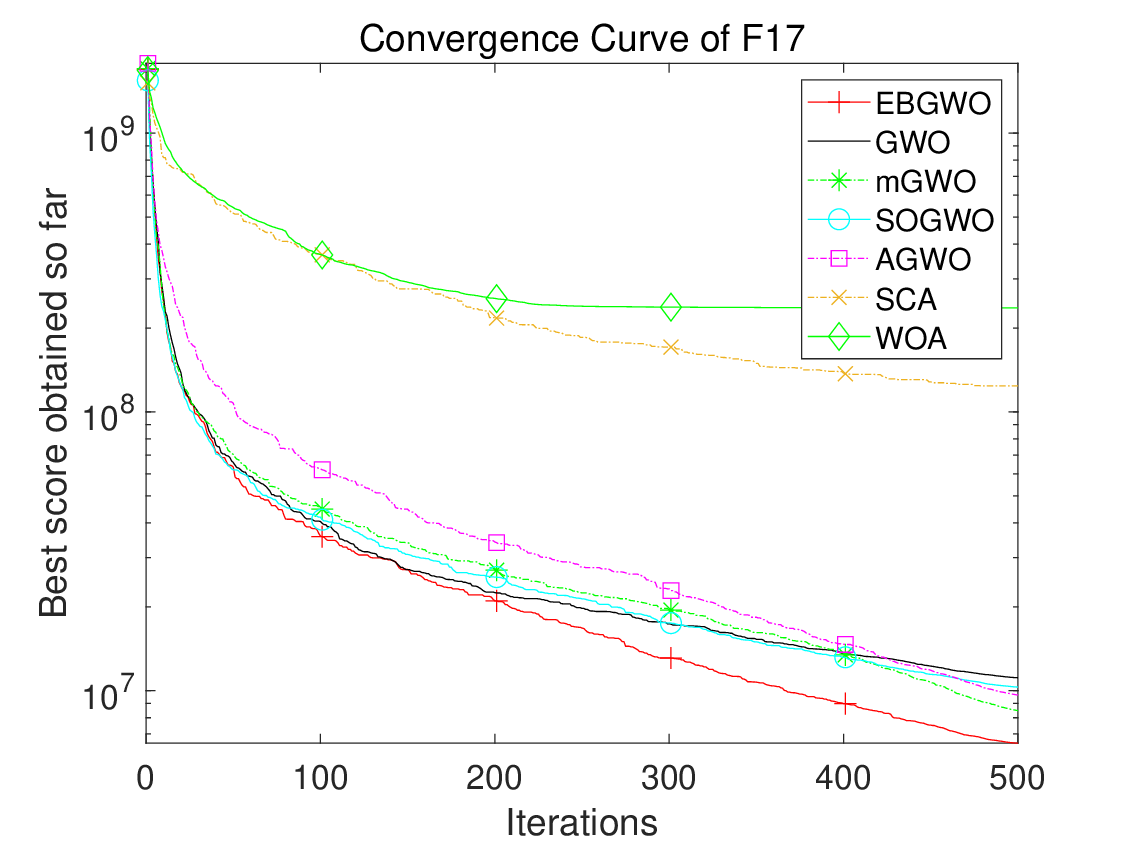}    
\label{cfig:50-3}}
\subfigure[$f_{30}$  ]{
\includegraphics[width=2in]{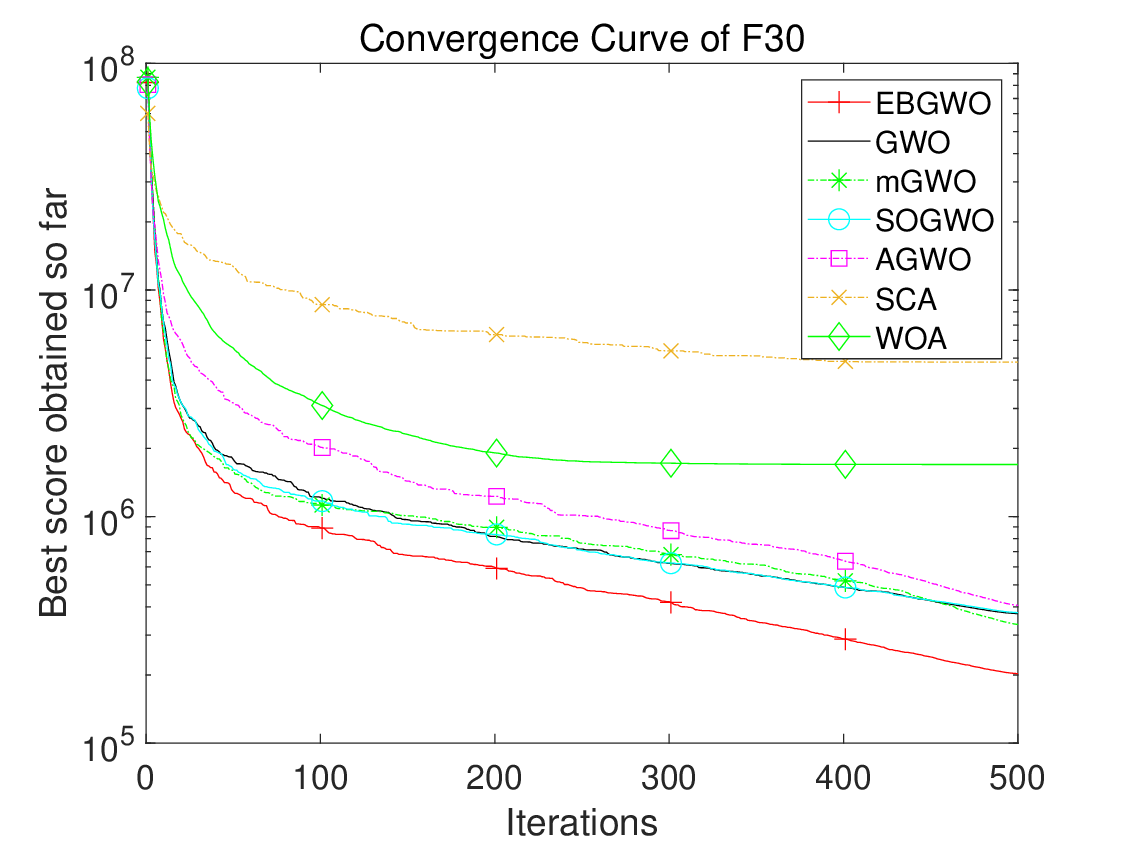}    
\label{cfig:50-4}}
\hspace{0.02\textwidth}
\caption{Convergence curves of EBGWO and other algorithms, Dim=50}
\label{cfig:50}
\end{figure}

\begin{figure}[ht]
\centering
\subfigure[$f_2$]{
\includegraphics[width=2in]{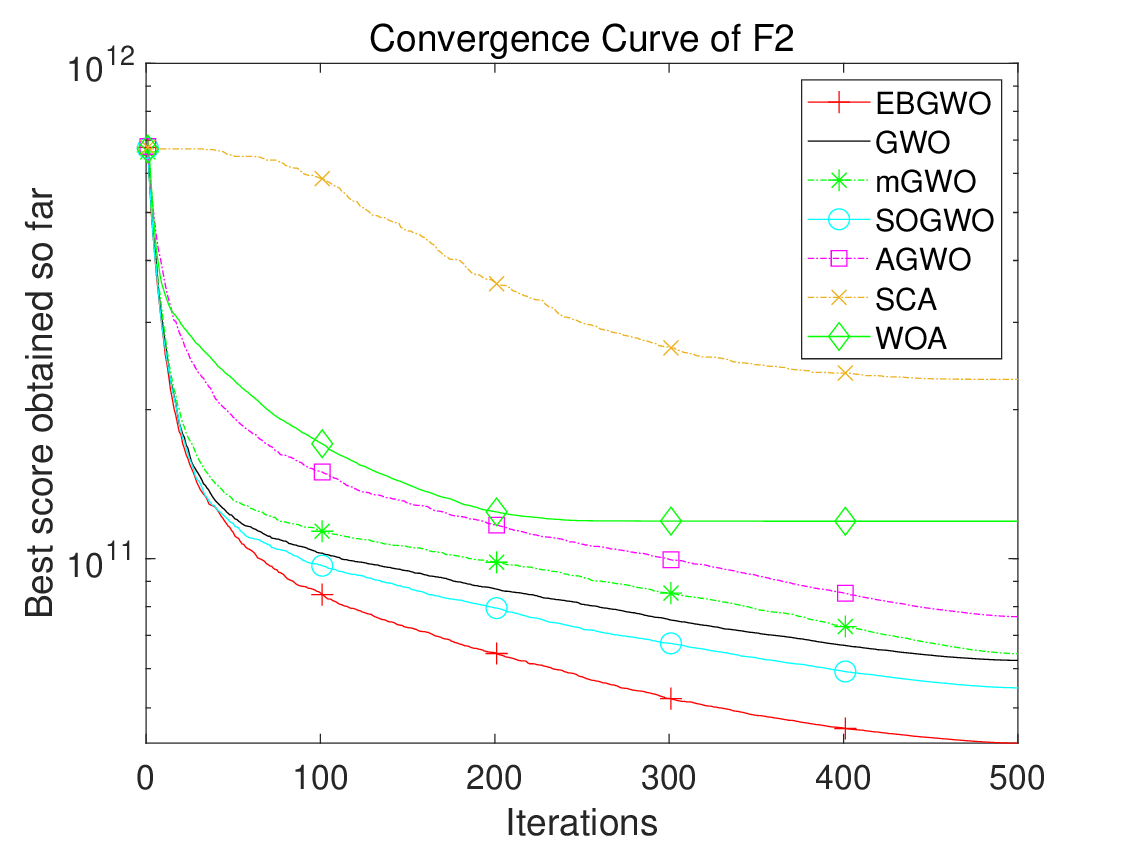}   
\label{cfig:100-1}}
\subfigure[$f_{11}$ ]{
\includegraphics[width=2in]{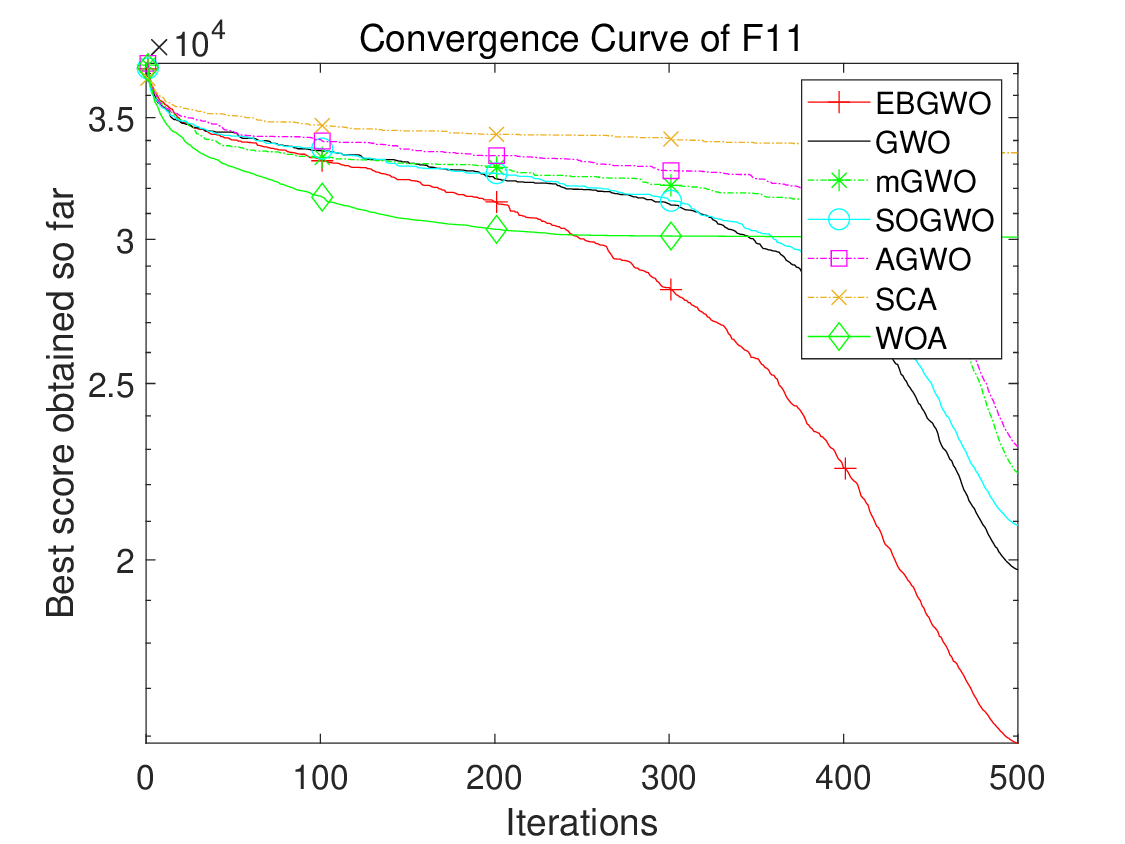}    
\label{cfig:100-2}}
\subfigure[$f_{21}$]{
\includegraphics[width=2in]{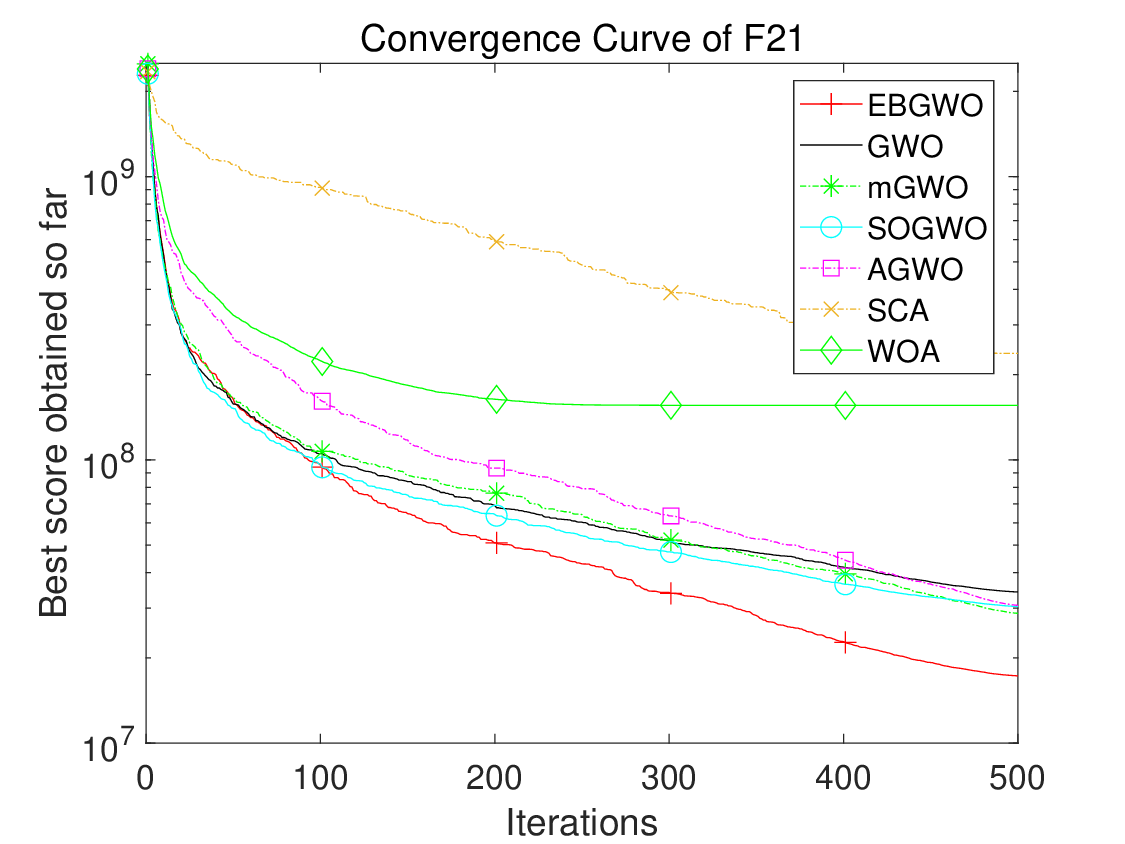}    
\label{cfig:100-3}}
\subfigure[$f_{28}$]{
\includegraphics[width=2in]{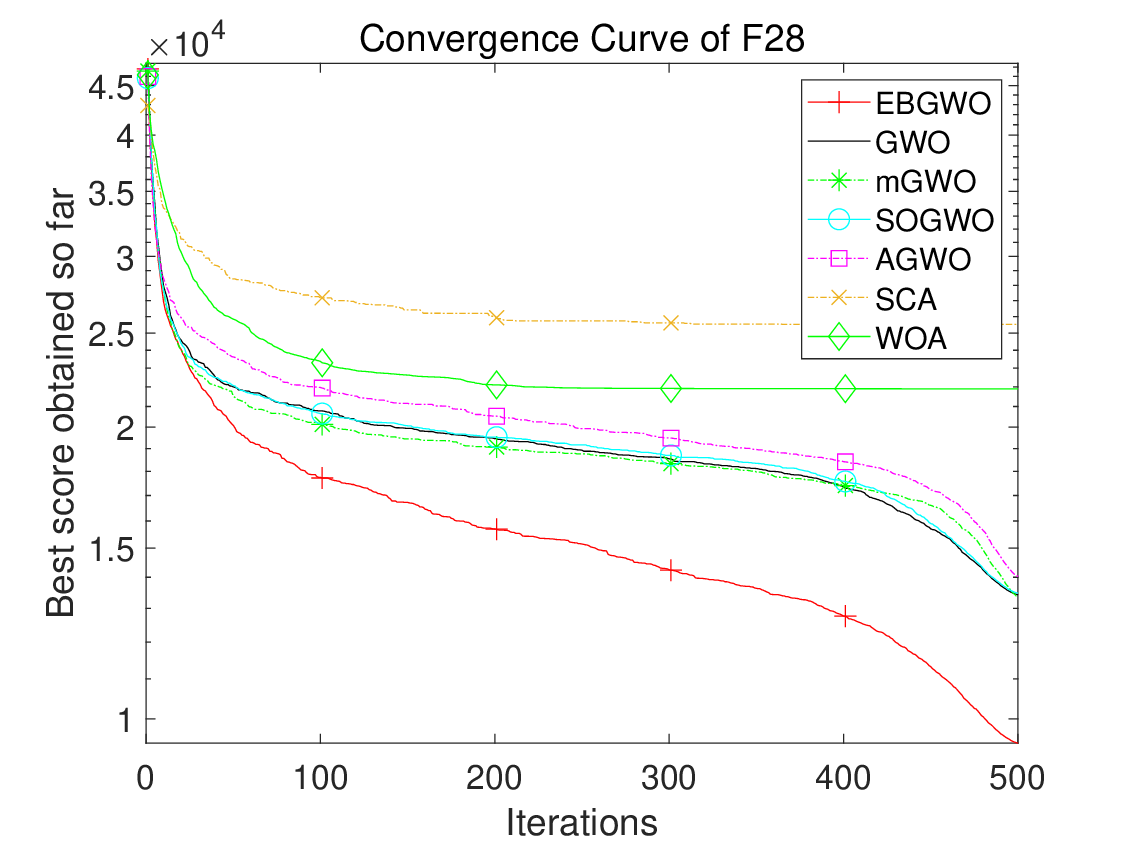}    
\label{cfig:100-4}}
\hspace{0.02\textwidth}
\caption{Convergence curves of EBGWO and other algorithms, Dim=100}
\label{cfig:100}
\end{figure}

\begin{figure}[ht]
\centering
\subfigure[$f_1$]{
\includegraphics[width=2in]{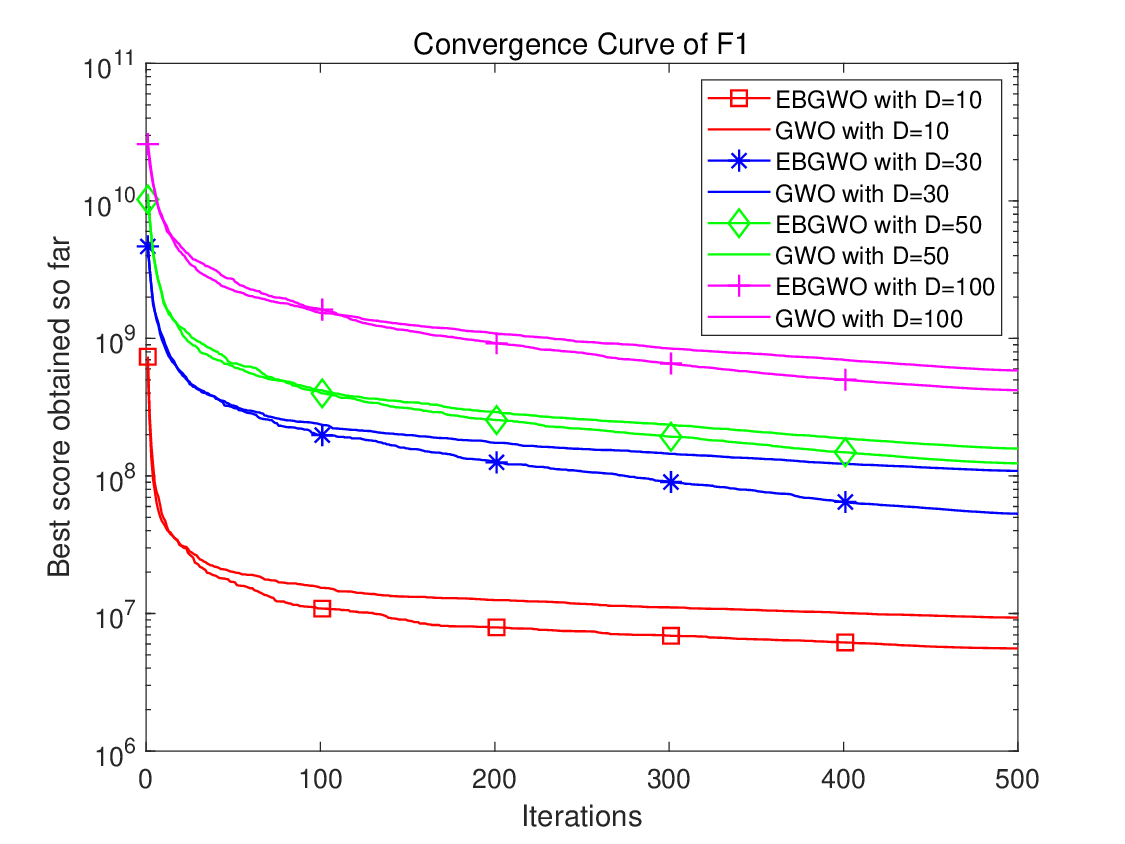}   
\label{DDfig:1}}
\subfigure[$f_{10}$ ]{
\includegraphics[width=2in]{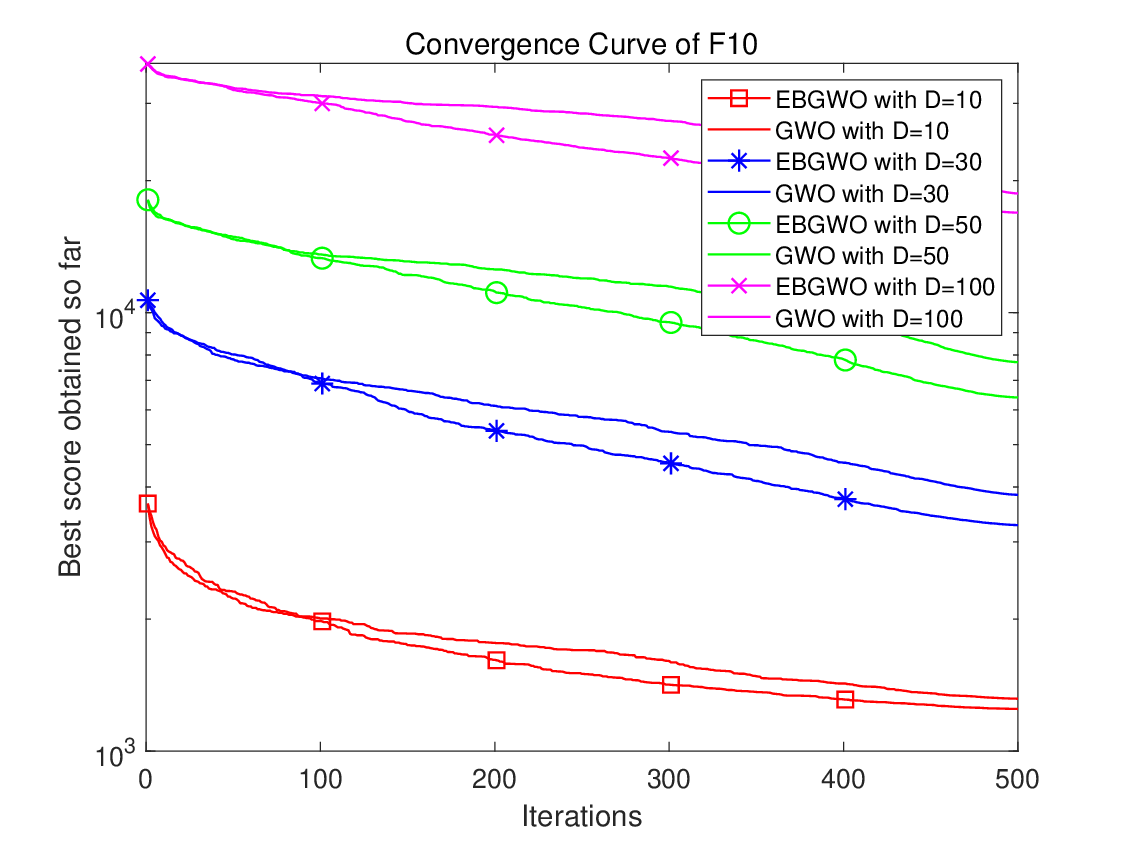}    
\label{DDfig:2}}
\subfigure[$f_{17}$]{
\includegraphics[width=2in]{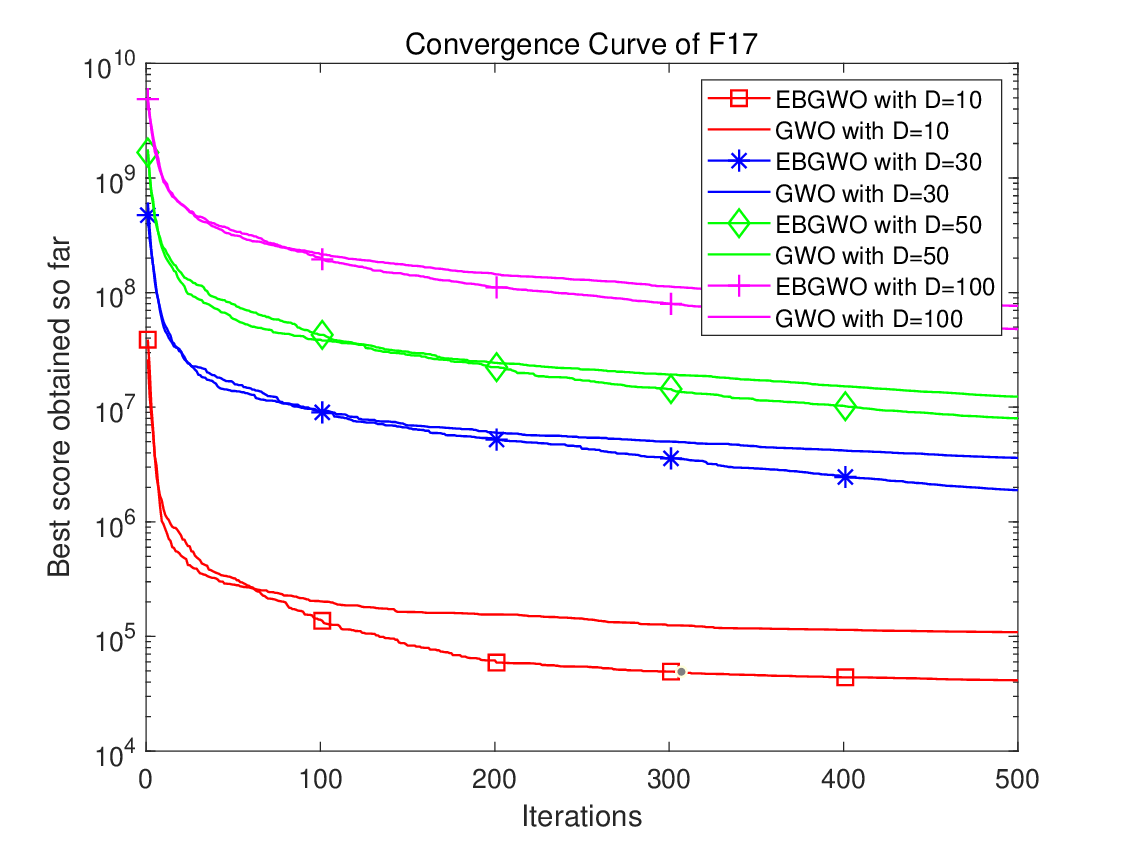}    
\label{DDfig:3}}
\subfigure[$f_{29}$]{
\includegraphics[width=2in]{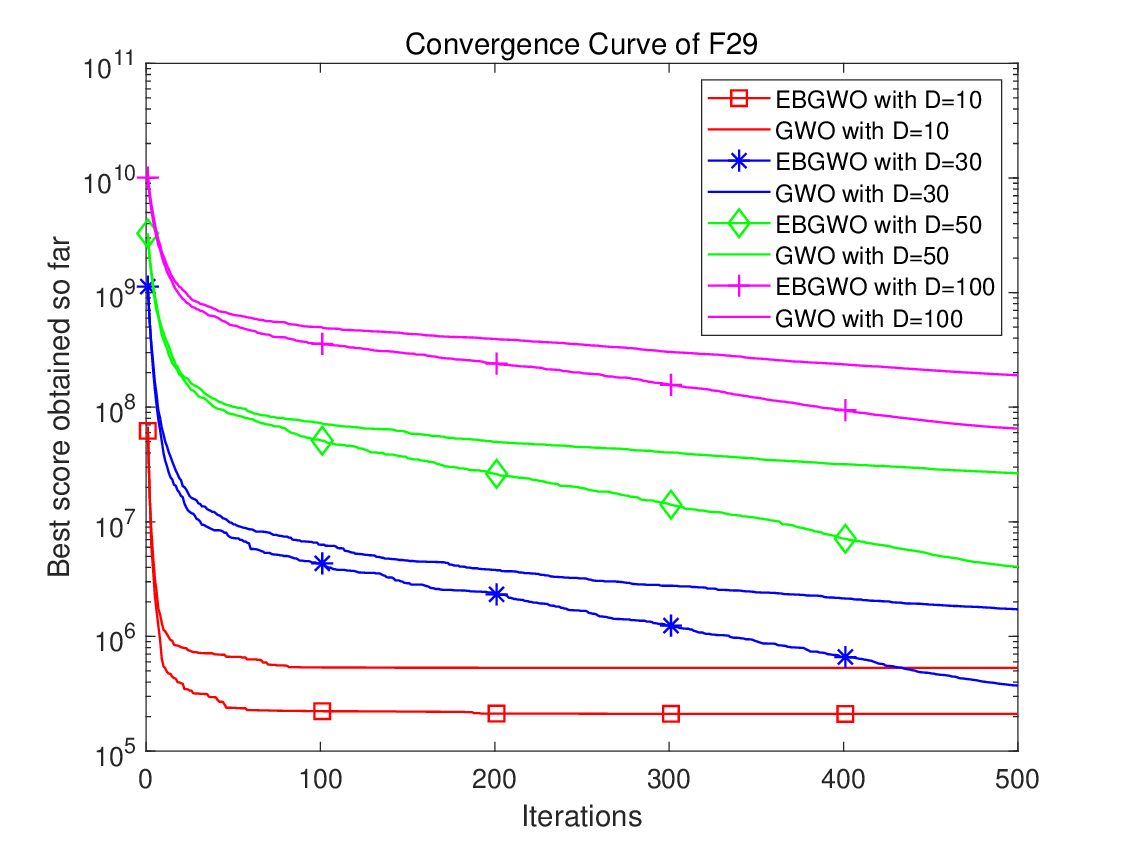}    
\label{DDfig:4}}
\hspace{0.02\textwidth}
\caption{Convergence curves of EBGWO in four different dimensions}
\label{DDfig}
\end{figure}

The quantitative analysis includes exploration and exploitation analysis, local optimum avoidance analysis, and convergence behaviour analysis. The mean, standard deviation, and ranking experimental results of baselines on benchmark functions in different dimensions are shown in Tables \ref{other1} to \ref{other4}. To highlight the differences, the best values in the table are marked in bold. Figs. \ref{cfig:10} to \ref{cfig:100} illustrate the convergence curve of EBGWO and baselines in four dimensions. Figure \ref{DDfig} presents convergence curves of the GWO and EBGWO algorithm in four different dimensions.

The proposed EBGWO algorithm is observed to obtain the best solutions for unimodal functions in four dimensions. As for the unimodal functions in Figs. \ref{cfig:10-1} to \ref{cfig:100-1}, the EBGWO curve appears lower than other curves, and the descent speed is faster than others. Given the characteristics of unimodal functions, the EBGWO algorithm demonstrates a strong exploitation capability. Meantime, for the multimodal functions in Figs. \ref{cfig:10-2} to \ref{cfig:100-2}, EBGWO performs better than other algorithms due to its good exploration capability. The balance search mechanism employed by the EBGWO algorithm expands the search region, thereby enhancing the possibility of finding the optimal solution. In terms of hybrid and composite functions presented in Figs. \ref{cfig:10-3}, \ref{cfig:10-4}, \ref{cfig:30-3}, \ref{cfig:30-4}, \ref{cfig:50-3}, \ref{cfig:50-4}, \ref{cfig:100-3} and \ref{cfig:100-4}, the convergence curves indicate that the EBGWO algorithm has a stronger exploration capability. It can better avoid the local optimum in the benchmark functions suite experiment. The remaining (w/t/l) values suggest that the EBGWO algorithm has achieved significant achievements compared to previous GWO variants. 

As shown in Table \ref{OE2}, the average and total ranking of EBGWO perform the best among baselines in the (w/t/l) rank calculation.

In summary, the elite inheritance mechanism enhances optimization accuracy, while the balance search mechanism achieves an optimal balance between exploration and exploitation through the use of a new operator and a novel selection of leading wolves. The elite inheritance mechanism and balance search mechanism introduced in EBGWO provide a more effective way to generate new positions. 

\begin{table}[htbp]
\begin{center}
  \caption{The Overall Effectiveness of EBGWO and other algorithms}  \label{OE2}
        \tabcolsep 7pt 
  \resizebox{4.8in}{15mm}{
\begin{tabular}{cccccccc}
    \hline
          & EBGWO & GWO   & mGWO  & SOGWO & AGWO  & SCA   & WOA \\
          & (w/t/l) & (w/t/l) & (w/t/l) & (w/t/l) & (w/t/l) & (w/t/l) & (w/t/l) \\
    \hline
    Dim=10  & (19/0/11) & (1/0/29) & (0/0/30) & (3/0/27) & (4/0/26) & (1/0/29) & (2/0/28) \\\hline
    Dim=30  & (26/0/4) & (0/0/30) & (0/0/30) & (0/0/30) & (1/0/29) & (1/0/29) & (2/0/28) \\\hline
    Dim=50  & (25/0/5) & (0/0/30) & (0/0/30) & (0/0/30) & (2/0/28) & (1/0/29) & (2/0/28) \\\hline
    Dim=100 & (25/0/5) & (0/0/30) & (1/0/29) & (0/0/30) & (0/0/30) & (0/0/30) & (4/0/26) \\\hline
    Total & (95/0/25) & (1/0/119) & (1/0/119) & (3/0/117) & (7/0/113) & (3/0/117) & (10/0/110) \\\hline
    OE    & \textbf{79.17\%} & 0.83\% & 0.83\% & 2.50\% & 5.83\% & 2.50\% & 8.33\% \\
\hline
    \end{tabular}}
    \end{center}
\end{table}%

\begin{table}[htbp]
\begin{center}
    \caption{The comparison of obtained solutions for unimodal functions ($F1-F3$)}  \label{other1}
  \resizebox{4.8in}{40mm}{
    \begin{tabular}{cccccccccc}
    \hline
    F     & D     & Index & EBGWO & GWO   & mGWO  & SOGWO & AGWO  & SCA   & WOA \\
    \hline
    \multirow{8}[2]{*}{F1} & 10    & Mean  & \multicolumn{1}{r}{6.41980E+06} & \multicolumn{1}{r}{7.85742E+06} & \multicolumn{1}{r}{7.72783E+06} & \multicolumn{1}{r}{6.98537E+06} & \multicolumn{1}{r}{\textbf{6.37087E+06}} & \multicolumn{1}{r}{1.31329E+07} & \multicolumn{1}{r}{1.41846E+07} \\
          &       & Std   & \multicolumn{1}{r}{5.63407E+06} & \multicolumn{1}{r}{4.53010E+06} & \multicolumn{1}{r}{6.06843E+06} & \multicolumn{1}{r}{4.93933E+06} & \multicolumn{1}{r}{6.56781E+06} & \multicolumn{1}{r}{5.39796E+06} & \multicolumn{1}{r}{1.10301E+07} \\
          & 30    & Mean  & \multicolumn{1}{r}{\textbf{5.98630E+07}} & \multicolumn{1}{r}{1.05757E+08} & \multicolumn{1}{r}{9.42224E+07} & \multicolumn{1}{r}{1.23591E+08} & \multicolumn{1}{r}{8.02140E+07} & \multicolumn{1}{r}{4.67482E+08} & \multicolumn{1}{r}{2.61992E+08} \\
          &       & Std   & \multicolumn{1}{r}{3.68459E+07} & \multicolumn{1}{r}{6.33026E+07} & \multicolumn{1}{r}{5.05473E+07} & \multicolumn{1}{r}{6.76688E+07} & \multicolumn{1}{r}{4.22417E+07} & \multicolumn{1}{r}{1.22066E+08} & \multicolumn{1}{r}{1.19043E+08} \\
          & 50    & Mean  & \multicolumn{1}{r}{\textbf{1.01889E+08}} & \multicolumn{1}{r}{1.42716E+08} & \multicolumn{1}{r}{1.78326E+08} & \multicolumn{1}{r}{1.66469E+08} & \multicolumn{1}{r}{1.36596E+08} & \multicolumn{1}{r}{1.26579E+09} & \multicolumn{1}{r}{4.34805E+08} \\
          &       & Std   & \multicolumn{1}{r}{5.26585E+07} & \multicolumn{1}{r}{5.40798E+07} & \multicolumn{1}{r}{7.25269E+07} & \multicolumn{1}{r}{7.89791E+07} & \multicolumn{1}{r}{5.73559E+07} & \multicolumn{1}{r}{4.28844E+08} & \multicolumn{1}{r}{1.24420E+08} \\
\hline
\end{tabular}}
\end{center}
\end{table}%

\begin{table}[htbp]
\begin{center}
  \caption{The comparison of obtained solutions for multimodal functions ($F_{4}-F_{16}$)}  \label{other2}
  \resizebox{4.8in}{93mm}{
    \begin{tabular}{cccccccccc}
    \hline
    F     & D     & Index & EBGWO & GWO   & mGWO  & SOGWO & AGWO  & SCA   & WOA \\
    \hline
    \multirow{8}[2]{*}{F4} & 10    & Mean  & \multicolumn{1}{r}{\textbf{4.32313E+02}} & \multicolumn{1}{r}{4.32393E+02} & \multicolumn{1}{r}{4.34116E+02} & \multicolumn{1}{r}{4.35286E+02} & \multicolumn{1}{r}{4.32747E+02} & \multicolumn{1}{r}{4.83393E+02} & \multicolumn{1}{r}{4.45172E+02} \\
          &       & Std   & \multicolumn{1}{r}{8.13101E+00} & \multicolumn{1}{r}{8.40071E+00} & \multicolumn{1}{r}{7.95685E+00} & \multicolumn{1}{r}{5.53819E+00} & \multicolumn{1}{r}{8.93239E+00} & \multicolumn{1}{r}{3.14111E+01} & \multicolumn{1}{r}{2.45389E+01} \\
          & 30    & Mean  & \multicolumn{1}{r}{\textbf{6.32660E+02}} & \multicolumn{1}{r}{7.60191E+02} & \multicolumn{1}{r}{7.25970E+02} & \multicolumn{1}{r}{7.13766E+02} & \multicolumn{1}{r}{7.68682E+02} & \multicolumn{1}{r}{3.05906E+03} & \multicolumn{1}{r}{1.27591E+03} \\
          &       & Std   & \multicolumn{1}{r}{8.08572E+01} & \multicolumn{1}{r}{2.94059E+02} & \multicolumn{1}{r}{1.30153E+02} & \multicolumn{1}{r}{9.71026E+01} & \multicolumn{1}{r}{2.00249E+02} & \multicolumn{1}{r}{7.63873E+02} & \multicolumn{1}{r}{3.48939E+02} \\
          & 50    & Mean  & \multicolumn{1}{r}{\textbf{1.11031E+03}} & \multicolumn{1}{r}{1.83044E+03} & \multicolumn{1}{r}{1.80874E+03} & \multicolumn{1}{r}{1.80702E+03} & \multicolumn{1}{r}{2.38664E+03} & \multicolumn{1}{r}{1.62623E+04} & \multicolumn{1}{r}{4.55504E+03} \\
          &       & Std   & \multicolumn{1}{r}{2.49329E+02} & \multicolumn{1}{r}{6.25380E+02} & \multicolumn{1}{r}{6.75762E+02} & \multicolumn{1}{r}{6.21684E+02} & \multicolumn{1}{r}{8.84883E+02} & \multicolumn{1}{r}{2.96285E+03} & \multicolumn{1}{r}{1.19368E+03} \\
\hline
    \end{tabular}}
    \end{center}
\end{table}%

\begin{table}[htbp]
\begin{center}
 \caption{The comparison of obtained solutions for hybrid functions ($F_{17}-F_{22}$)} \label{other3} 
 \resizebox{4.8in}{75mm}{
    \begin{tabular}{cccccccccc}
    \hline
    F     & D     & Index & EBGWO & GWO   & mGWO  & SOGWO & AGWO  & SCA   & WOA \\
    \hline
    \multirow{8}[2]{*}{F17} & 10    & Mean  & \multicolumn{1}{r}{\textbf{1.74953E+04}} & \multicolumn{1}{r}{8.49406E+04} & \multicolumn{1}{r}{3.53460E+04} & \multicolumn{1}{r}{1.89220E+05} & \multicolumn{1}{r}{1.95722E+04} & \multicolumn{1}{r}{6.63798E+04} & \multicolumn{1}{r}{2.25775E+05} \\
          &       & Std   & \multicolumn{1}{r}{6.13063E+04} & \multicolumn{1}{r}{1.64204E+05} & \multicolumn{1}{r}{8.89211E+04} & \multicolumn{1}{r}{2.27616E+05} & \multicolumn{1}{r}{6.42574E+04} & \multicolumn{1}{r}{8.43401E+04} & \multicolumn{1}{r}{3.58397E+05} \\
          & 30    & Mean  & \multicolumn{1}{r}{\textbf{2.26791E+06}} & \multicolumn{1}{r}{3.24444E+06} & \multicolumn{1}{r}{3.74272E+06} & \multicolumn{1}{r}{5.17468E+06} & \multicolumn{1}{r}{3.55736E+06} & \multicolumn{1}{r}{2.08446E+07} & \multicolumn{1}{r}{3.06380E+07} \\
          &       & Std   & \multicolumn{1}{r}{2.16582E+06} & \multicolumn{1}{r}{3.57471E+06} & \multicolumn{1}{r}{2.76557E+06} & \multicolumn{1}{r}{4.98617E+06} & \multicolumn{1}{r}{3.34007E+06} & \multicolumn{1}{r}{1.23458E+07} & \multicolumn{1}{r}{2.45321E+07} \\
          & 50    & Mean  & \multicolumn{1}{r}{\textbf{6.48240E+06}} & \multicolumn{1}{r}{1.11310E+07} & \multicolumn{1}{r}{8.48426E+06} & \multicolumn{1}{r}{1.03088E+07} & \multicolumn{1}{r}{9.65620E+06} & \multicolumn{1}{r}{1.24005E+08} & \multicolumn{1}{r}{2.36160E+08} \\
          &       & Std   & \multicolumn{1}{r}{5.35971E+06} & \multicolumn{1}{r}{6.84564E+06} & \multicolumn{1}{r}{6.20355E+06} & \multicolumn{1}{r}{8.03067E+06} & \multicolumn{1}{r}{6.35576E+06} & \multicolumn{1}{r}{3.93202E+07} & \multicolumn{1}{r}{1.24014E+08} \\
\hline
    \end{tabular}}
    \end{center}
\end{table}%

\begin{table}[htbp]
\begin{center}
\caption{The comparison of obtained solutions for composite functions ($F_{23}-F_{30}$)} \label{other4} 
\resizebox{4.8in}{80mm}{
    \begin{tabular}{cccccccccc}
    \hline
    F     & D     & Index & EBGWO & GWO   & mGWO  & SOGWO & AGWO  & SCA   & WOA \\
    \hline
    \multirow{8}[2]{*}{F23} & 10    & Mean  & \multicolumn{1}{r}{2.63199E+03} & \multicolumn{1}{r}{2.63424E+03} & \multicolumn{1}{r}{2.63605E+03} & \multicolumn{1}{r}{2.63475E+03} & \multicolumn{1}{r}{2.63496E+03} & \multicolumn{1}{r}{2.64740E+03} & \multicolumn{1}{r}{\textbf{2.63174E+03}} \\
          &       & Std   & \multicolumn{1}{r}{2.37178E+00} & \multicolumn{1}{r}{4.53507E+00} & \multicolumn{1}{r}{6.03668E+00} & \multicolumn{1}{r}{4.20893E+00} & \multicolumn{1}{r}{3.45440E+00} & \multicolumn{1}{r}{5.34248E+00} & \multicolumn{1}{r}{3.71015E+01} \\
          & 30    & Mean  & \multicolumn{1}{r}{\textbf{2.63373E+03}} & \multicolumn{1}{r}{2.64235E+03} & \multicolumn{1}{r}{2.64187E+03} & \multicolumn{1}{r}{2.64072E+03} & \multicolumn{1}{r}{2.64346E+03} & \multicolumn{1}{r}{2.72651E+03} & \multicolumn{1}{r}{2.71014E+03} \\
          &       & Std   & \multicolumn{1}{r}{5.94789E+00} & \multicolumn{1}{r}{1.06707E+01} & \multicolumn{1}{r}{1.15163E+01} & \multicolumn{1}{r}{1.38116E+01} & \multicolumn{1}{r}{1.28813E+01} & \multicolumn{1}{r}{3.18128E+01} & \multicolumn{1}{r}{2.86062E+01} \\
          & 50    & Mean  & \multicolumn{1}{r}{\textbf{2.74121E+03}} & \multicolumn{1}{r}{2.80173E+03} & \multicolumn{1}{r}{2.79683E+03} & \multicolumn{1}{r}{2.79406E+03} & \multicolumn{1}{r}{2.82254E+03} & \multicolumn{1}{r}{3.24697E+03} & \multicolumn{1}{r}{2.85887E+03} \\
          &       & Std   & \multicolumn{1}{r}{2.81753E+01} & \multicolumn{1}{r}{4.99947E+01} & \multicolumn{1}{r}{7.21562E+01} & \multicolumn{1}{r}{7.06433E+01} & \multicolumn{1}{r}{5.28097E+01} & \multicolumn{1}{r}{1.24300E+02} & \multicolumn{1}{r}{2.15421E+02} \\
\hline
    \end{tabular}}
    \end{center}
\end{table}%

\subsubsection{Ablation Study}
To more accurately assess the impact of each mechanism embedded in the proposed algorithm, an ablation study is conducted in this subsection. Three versions of the algorithm are evaluated: the GWO algorithm with elite inheritance mechanism, the GWO algorithm with balance search mechanism, and the proposed EBGWO algorithm that integrates both mechanisms. This study utilizes the same data sets previously mentioned, specifically the classic IEEE CEC 2014 benchmark functions. Table \ref{ablation} presents a comparative analysis of the number of optimal solutions (wins), identical solutions (ties), and suboptimal solutions (losses) derived from testing each algorithm variant against the traditional Grey Wolf Optimizer (GWO) algorithm.

As illustrated in Table \ref{ablation}, the integration of both mechanisms substantially improves the overall efficacy of the algorithm, indicating their collective importance to the final results. Interestingly, we observe that the use of any single mechanism within the proposed algorithm does not necessarily outperform the classical GWO algorithm. This could be attributed to the potential of the elite inheritance mechanism to weaken exploration performance and increase the risk of local optimum convergence, despite its ability to enhance the accuracy of guidance information and exploitation performance. However, the introduction of the balance search mechanism can mitigate this risk by facilitating escape from local optima and enhancing exploration efficiency. Meanwhile, the effective implementation of the balance search mechanism is contingent upon the elite inheritance mechanism, as there is an implicit interdependence between them. Through the inheritance of elite positions, the search process can leverage more valuable information and explore more effectively.

In conclusion, the combination of these two mechanisms is crucial to augmenting the overall performance of the GWO algorithm.

\begin{table}[ht]
  \centering
  \caption{Ablation study: Effects of two mechanisms in EBGWO algorithm tested on IEEE CEC 2014 Benchmark Function Suite. EIM: Elite Inheritance Mechanism, BSM: Balance Search Mechanism}
  \resizebox{4.8in}{12mm}{
    \begin{tabular}{l|c|c|c|c}
    \hline
    \multicolumn{1}{l|}{Algorithm} & \multicolumn{4}{c}{(w/t/l)} \\
    \hline
    \multicolumn{1}{l|}{Classic GWO vs.} & Dim=10 & Dim=30 & Dim=50 & Dim=100 \\
    \hline
    GWO+ EIM & (3/0/27) & (7/0/23) & (5/0/25) & (6/0/24) \\
    \hline
    GWO+ BSM & (22/0/8) & (16/0/24) & (10/0/20) & (12/0/18) \\
    \hline
    GWO+ EIM + BSM (EBGWO) & \textbf{(26/0/4)} & \textbf{(28/0/2)} & \textbf{(28/0/2)} & \textbf{(29/0/1)} \\
    \hline
    \end{tabular}}%
  \label{ablation}%
\end{table}%

\subsection{Experiment 2: Statistical Testing}
The goal of Experiment 2 is to assess the significant differences between the EBGWO algorithm and other algorithms using the benchmark functions suite, utilizing the Wilcoxon rank-sum test for statistical analysis \citep{33}. Wilcoxon rank-sum test is a common nonparametric test used to determine whether two distribution columns are significantly different \citep{34}.

Two \textbf{hypotheses} are given in the paper and evaluated using Wilcoxon's test.
\begin{itemize}
\item $ H_{0} $-null hypothesis: The EBGWO algorithm with elite inheritance mechanism and balance search mechanism performs worse in benchmark tests compared with baselines.
\item $ H_{1} $-alternative hypothesis: The EBGWO algorithm with elite inheritance mechanism and balance search mechanism performs well in benchmark tests compared with baselines.
\end{itemize}

The results of the Wilcoxon test comparing the EBGWO algorithm with the baseline algorithms across four different dimensions are provided in Table \ref{Wilcoxon}. According to the experimental results, the alternative hypothesis ($H_{1}$) is supported. This supports that the EBGWO algorithm is more capable of finding global optimal solutions than other algorithms and can better maintain a balance between exploration and exploitation. 
\begin{table}[ht]
\caption{Results of the Wilcoxon’s test between EBGWO and other algorithms on IEEE CEC 2014 benchmark functions} \label{Wilcoxon} 
\centering 
      \tabcolsep 10pt 
\resizebox{4.8in}{55mm}{
    \begin{tabular}{ccccccccc}
    \hline
    Algorithms & \multicolumn{8}{c}{Dim=10} \\
    \hline
    EBGWO vs. & Better & Equal & Worst & $ W^{+} $    & $W^{-} $    & $p$-value & $\alpha$=0.05 & $\alpha$=0.1 \\
    \hline
    GWO   & 26    & 0     & 4     & 418   & 47    & 2.05153E-04 & Yes   & Yes \\
    mGWO  & 27    & 1     & 2     & 437   & 25    & 1.07963E-05 & Yes   & Yes \\
    SOGWO & 24    & 0     & 6     & 371   & 94    & 1.47728E-04 & Yes   & Yes \\
    AGWO  & 22    & 0     & 8     & 338   & 127   & 1.24526E-02 & Yes   & Yes \\
    SCA   & 27    & 0     & 3     & 430   & 35    & 4.86026E-05 & Yes   & Yes \\
    WOA   & 25    & 0     & 5     & 379   & 86    & 8.18775E-05 & Yes   & Yes \\
    \hline
    Algorithms & \multicolumn{8}{c}{Dim=30} \\
    \hline
    EBGWO vs. & Better & Equal & Worst & $ W^{+} $    & $W^{-} $    & $p$-value & $\alpha$=0.05 & $\alpha$=0.1 \\
    \hline
    GWO   & 28    & 0     & 2     & 436   & 29    & 4.72920E-06 & Yes   & Yes \\
    mGWO  & 29    & 0     & 1     & 439   & 26    & 2.35342E-06 & Yes   & Yes \\
    SOGWO & 28    & 0     & 2     & 424   & 41    & 6.98378E-06 & Yes   & Yes \\
    AGWO  & 27    & 0     & 3     & 408   & 57    & 5.30699E-05 & Yes   & Yes \\
    SCA   & 28    & 0     & 2     & 424   & 41    & 4.28569E-06 & Yes   & Yes \\
    WOA   & 28    & 0     & 2     & 424   & 41    & 1.97295E-05 & Yes   & Yes \\
    \hline
    Algorithms & \multicolumn{8}{c}{Dim=50} \\
    \hline
    EBGWO vs. & Better & Equal & Worst  & $ W^{+} $    & $W^{-} $    & $p$-value & $\alpha$=0.05 & $\alpha$=0.1 \\
    \hline
    GWO   & 29    & 0     & 1     & 453   & 12    & 1.92092E-06 & Yes   & Yes \\
    mGWO  & 29    & 0     & 1     & 449   & 16    & 2.12664E-06 & Yes   & Yes \\
    SOGWO & 29    & 0     & 1     & 453   & 12    & 2.12664E-06 & Yes   & Yes \\
    AGWO  & 26    & 0     & 4     & 410   & 55    & 1.35948E-04 & Yes   & Yes \\
    SCA   & 28    & 0     & 2     & 442   & 23    & 3.51524E-06 & Yes   & Yes \\
    WOA   & 27    & 0     & 3     & 390   & 75    & 7.69086E-06 & Yes   & Yes \\
    \hline
    Algorithms & \multicolumn{8}{c}{Dim=100} \\
    \hline
    EBGWO vs. & Better & Equal & Worst & $ W^{+} $    & $W^{-} $    & $p$-value & $\alpha$=0.05 & $\alpha$=0.1\\
    \hline
    GWO   & 29    & 0     & 1     & 452   & 13    & 1.92092E-06 & Yes   & Yes \\
    mGWO  & 27    & 0     & 3     & 393   & 72    & 1.36011E-05 & Yes   & Yes \\
    SOGWO & 29    & 0     & 1     & 449   & 16    & 1.92092E-06 & Yes   & Yes \\
    AGWO  & 28    & 0     & 2     & 423   & 42    & 2.35342E-06 & Yes   & Yes \\
    SCA   & 29    & 0     & 1     & 459   & 6     & 1.92092E-06 & Yes   & Yes \\
    WOA   & 26    & 1     & 3     & 396   & 48    & 8.84792E-06 & Yes   & Yes \\
    \hline
    \end{tabular}}
\end{table}%

\subsection{Experiment 3: Engineering Design Problems} 
In Experiment 3, to further evaluate the performance of the proposed EBGWO algorithm, we apply it to three real-world engineering design problems. The purpose of the experiment is to verify the practical applicability of the algorithm. 

\subsubsection{Gear train design problem}
\begin{figure}[htbp]
\centering
\includegraphics[width=4in]{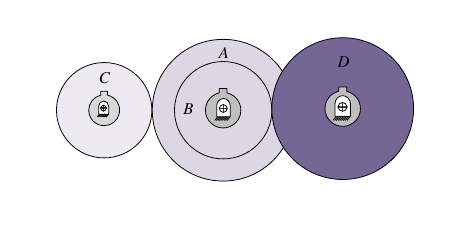}   
\caption{\label{GTDF}The Gear Train Design Problem}
\end{figure}

The gear train design problem is an unconstrained engineering problem. The objective of the problem is to minimize the error between a target gear ratio ($\dfrac{1}{6.931}$) and the actual obtained gear ratio. This problem involves four variables, each representing the number of teeth on a gear. The schematic diagram \citep{11,35,36} is shown in Fig. \ref{GTDF}. The problem can be characterized as follows:

Consider : $\overrightarrow{x}$=[$x_1$,$x_2$,$x_3$,$x_{4}$]=[$n_{A}$,$n_{B}$,$n_{C}$,$n_{D}$],

Minimize : $f(\overrightarrow{x})=(\dfrac{1}{6.931}-\dfrac{x_{3} \times x_{2}}{x_{1} \times x_{4}})^{2}$,

Subject to : $12 \leq x_1$,$x_2$,$x_3$,$x_{4} \leq 60$.

\begin{table}[htbp]
\begin{center}
  \caption{Comparison of the mean optimal solutions for gear train design problem} \label{res:GTD1}
      \tabcolsep 10pt 
\resizebox{4.8in}{20mm}{
    \begin{tabular}{cccccc}
    \hline
    \multirow{2}[2]{*}{Algorithms} & \multicolumn{4}{c}{Variables} & \multirow{2}[2]{*}{$f(x)$ } \\
       \cline{2-5}   &$x_{1}$    &$x_{2}$   & $x_{3}$   & $x_{4}$    &  \\
    \hline
    EBGWO & 44 & 19 & 16 & 46 & \textbf{5.96732E-10} \\
    GWO   & 50 & 21 & 18 & 49 & 4.60448E-08 \\
    mGWO  & 48 & 18 & 19 & 48 & 7.54186E-07 \\
    AGWO  & 49 & 19 & 20 & 50 & 5.67960E-07 \\
    SOGWO & 51 & 20 & 20 & 51 & 2.34993E-08 \\
    SCA   & 49 & 17 & 18 & 44 & 1.49189E-04 \\
    WOA   & 49 & 19 & 18 & 46 & 1.40822E-06 \\
    \hline
    \end{tabular}}
    \end{center}
\end{table}%

\begin{table}[htbp]
\begin{center}
  \caption{Statistical results of five methods for gear train design problem} \label{res:GTD2}
      \tabcolsep 12pt 
      \resizebox{4.8in}{15mm}{
    \begin{tabular}{ccccc}
    \hline
    Algorithms & Best  & Mean  & Worst & Standard Deviation \\
    \hline
    EBGWO & \textbf{2.70086E-12} & \textbf{5.96732E-10} & \textbf{2.35764E-09} & \textbf{6.17444E-10} \\
    GWO   & 9.93988E-11 & 4.60448E-08 & 4.77605E-07 & 9.83617E-08 \\
    mGWO  & \textbf{2.70086E-12} & 7.54186E-07 & 8.0657E-06 & 1.88753E-06 \\
    AGWO  & 9.93988E-11 & 5.6796E-07 & 9.63861E-06 & 2.1278E-06 \\
    SOGWO & \textbf{2.70086E-12} & 2.34993E-08 & 2.09322E-07 & 4.09296E-08 \\
    SCA   & 2.41494E-06 & 1.49189E-04 & 1.03336E-03 & 2.12181E-04 \\
    WOA   & 8.88761E-10 & 1.40822E-06 & 2.60410E-05 & 5.06818E-06 \\
    \hline
    \end{tabular}}
    \end{center}
\end{table}%

This paper applies EBGWO and baselines to solve the gear train design problem. The comparison results of the mean optimal solutions of seven algorithms are shown in Table \ref{res:GTD1}, and the statistical results are presented in Table \ref{res:GTD2}. The EBGWO, mGWO, and AGWO algorithms achieve the best values, with the EBGWO algorithm obtaining the lowest mean gear train ratio error compared to other algorithms over the same number of iterations. Furthermore, the statistical results indicate that the EBGWO algorithm performs better in this problem. It has small standard deviation, Mean, Best, and Worst values simultaneously.

\subsubsection{Pressure vessel design problem}
\begin{figure}[htbp]
\centering
\includegraphics[width=4in]{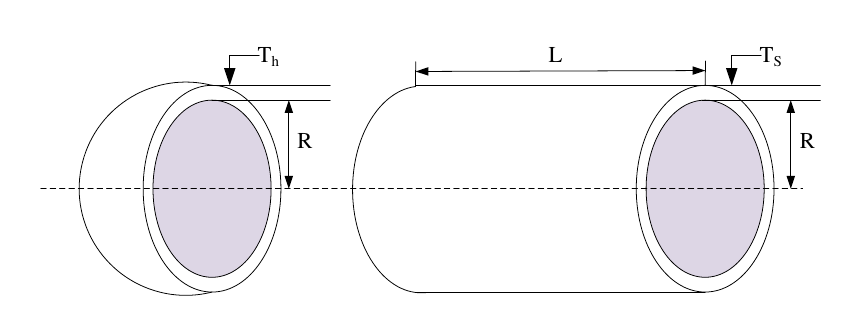}   
\caption{The Pressure Vessel Design Problem}
\label{fig:PVD}
\end{figure}

The pressure vessel design problem is demonstrated in Fig. \ref{fig:PVD}, which aims to minimize the cost of material, forming, and welding of a vessel \citep{37,38}. The problem has four variables including the thickness of the shell $T_{s}$ ($x_1$), the thickness of the head $T_{h}$ ($x_2$), the inner radius $R$ ($x_3$), and the length of the cylindrical section of the vessel, not including the head $L$ ($x_4$) \citep{39}. The problem can be described as follows:

Consider : $\overrightarrow{x}$=[$x_1$ $x_2$ $x_3$ $x_{4}$]=[$T_{s}$ $T_{h}$ $R$ $L$],\\

Minimize : $f(x)=0.6224x_1x_3x_4+1.7781x_2x_3^2+3.1661x_1^2x_4+19.84x_1^2x_3$,\\

Subject to : $g_1(x)=-x_1 +0.0193x_3 \leq 0$,\\

$g_2(x)=-x_2+0.00954 \leq 0$,\\

$g_3(x)= - \pi x_3^2 x_4- \dfrac{4}{3}\pi x_3^3 +1296000 \leq 0$,\\

$g_4(x)= x_4 -240 \leq 0$,\\

where : $0\leq x_{i}\leq 100, i=1, 2$,\\

$10\leq x_{i}\leq 200, i=3, 4$\\

\begin{table}[htbp]
\begin{center}
  \caption{Comparison of the mean optimal solutions for pressure vessel design problem} \label{res:PVD1}
      \tabcolsep 2pt 
      \resizebox{4.8in}{15mm}{
    \begin{tabular}{cccccccccc}
    \toprule
    \multirow{2}[4]{*}{Algorithms} & \multicolumn{8}{c}{Variables}                                 & \multirow{2}[4]{*}{$f(x)$} \\
\cmidrule{2-9}          & $x_1$    & $x_2$    & $x_3$    & $x_4$    & $g_1$    & $g_2$    & $g_3$    & $g_4$    &  \\
    \midrule
    EBGWO & 1.4E+01 & 5.8E+00 & 4.5E+01 & 1.6E+02 & 1.4E-02 & 6.6E-02 & -1.4E+02 & -8.1E+01 & \textbf{6.30933E+03} \\
    GWO   & 1.9E+01 & 2.5E+00 & 6.2E+01 & 2.7E+01 & -6.3E-03 & 4.3E-01 & -3.1E+03 & -2.1E+02 & 8.90879E+03 \\
    mGWO  & 1.8E+01 & 2.9E+00 & 5.7E+01 & 6.2E+01 & -7.8E-03 & 3.6E-01 & -5.5E+03 & -1.8E+02 & 9.12266E+03 \\
    AGWO  & 1.7E+01 & 3.1E+00 & 5.6E+01 & 7.1E+01 & -3.8E-03 & 3.4E-01 & -2.2E+03 & -1.7E+02 & 9.26121E+03 \\
    SOGWO & 1.7E+01 & 3.4E+00 & 5.6E+01 & 7.6E+01 & -3.6E-03 & 3.1E-01 & -1.7E+03 & -1.6E+02 & 8.30733E+03 \\
    SCA   & 1.7E+01 & 6.1E+00 & 5.7E+01 & 1.0E+02 & 1.6E-02 & 1.7E-01 & -4.3E+05 & -1.4E+02 & 1.04022E+04 \\
    WOA   & 1.6E+01 & 4.7E+00 & 5.5E+01 & 7.7E+01 & 1.8E-02 & 2.2E-01 & -5.3E+02 & -1.6E+02 & 9.12611E+03 \\
    \bottomrule
    \end{tabular}}%
        \end{center}
\end{table}%

\begin{table}[htbp]
\begin{center}
  \caption{Statistical results of five methods for pressure vessel design problem} \label{res:PVD2}
  \resizebox{4.8in}{15mm}{
      \tabcolsep 12pt 
    \begin{tabular}{ccccc}
    \hline
    Algorithms & Best  & Mean  & Worst & Standard Deviation \\
    \hline
    EBGWO & \textbf{6.05987E+03} & \textbf{6.30933E+03} & \textbf{7.58221E+03} & \textbf{5.46188E+02 }\\
    GWO   & 6.08000E+03 & 8.90879E+03 & 1.71586E+04 & 2.62677E+03 \\
    mGWO  & 6.08346E+03 & 9.12266E+03 & 2.33424E+04 & 4.15174E+03 \\
    AGWO  & 6.08435E+03 & 9.26121E+03 & 2.10711E+04 & 4.16561E+03 \\
    SOGWO & 6.06439E+03 & 8.30733E+03 & 1.80416E+04 & 2.97576E+03 \\
    SCA   & 6.94810E+03 & 1.04022E+04 & 1.46086E+04 & 2.23553E+03 \\
    WOA   & 6.11208E+03 & 9.12611E+03 & 3.53656E+04 & 5.40935E+03 \\
    \hline
    \end{tabular}}
    \end{center}
\end{table}%

Seven comparison algorithms are used to tackle this problem, and the performance of the EBGWO algorithm is assessed by comparing the optimal values it obtains with those from the other algorithms. The results are presented in Tables \ref{res:PVD1} and \ref{res:PVD2}. It is clear that the EBGWO has the most promising results compared with the baselines in this problem. On the other hand, the EBGWO algorithm gains the best statistical results among seven algorithms.

\subsubsection{Welded beam design problem}
\begin{figure}[htbp]
\centering
\includegraphics[width=4in]{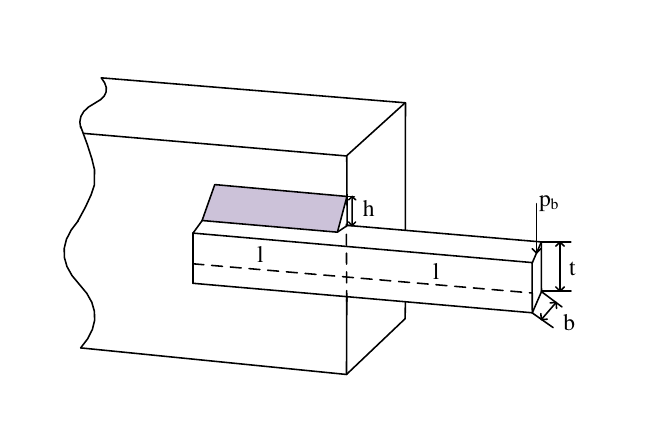}   
\caption{\label{wbd}The Welded Beam Design Problem}
\end{figure}

The welded beam design problem is a well-known problem in the field of structural optimization, which aims to minimize the fabrication cost of a welded beam \citep{3, 40}. The problem is shown in Fig. \ref{wbd}, and the constraints are as follows:

Consider : $\overrightarrow{x}$=[$x_1$ $x_2$ $x_3$ $x_{4}$]=[$h$ $l$ $t$ $b$],\\

Minimize: $F(x)=1.10471x_1^2x_2+0.04811x_3x_4(14.0+x_2)$,\\

Subject to :

$g_1(x)=\tau(x)-\tau_{max} \leq 0$,\\

$g_2(x)=\sigma(x)-\sigma_{max} \leq 0$,\\

$g_3(x)=\delta(x)-\delta_{max} \leq 0$,\\

$g_4(x)=x-1-x_4 \leq 0$,\\

$g_5(x)=P - P_c(x) \leq 0$,\\

$g_6(x)=0.125-x_1 \leq 0$,\\

$g_7(x)=0.10471x_1^2+0.04811x_3x_4(14.0+x_2)-5.0 \leq 0$,\\

where: $0.1 \leq x_{i}\leq 2, i=1, 2$,\\

$0.1\leq x_{i}\leq 10, i=3, 4$\\

\begin{table}[htbp]
\begin{center}
  \caption{Comparison of the mean optimal solutions for welded beam design problem} \label{res:WBD1}
  \resizebox{4.8in}{15mm}{
      \tabcolsep 2.5pt 
    \begin{tabular}{ccccccccccccc}
    \toprule
    \multirow{2}[4]{*}{Algorithms} & \multicolumn{11}{c}{Variables}                                                        & \multirow{2}[4]{*}{$f(x)$} \\
\cmidrule{2-12}          & $x_1$    & $x_2$    & $x_3$    & $x_4$    & $g_1$    & $g_2$    & $g_3$    & $g_4$    & $g_5$    & $g_6$    & $g_7$    &  \\
    \midrule
    EBGWO & 0.20  & 3.49  & 9.04  & 0.21  & -15.98  & -41.42  & 0.00  & -3.39  & -0.08  & -0.24  & -15.85  & \textbf{1.72871E+00} \\
    GWO   & 0.20  & 3.51  & 9.04  & 0.21  & -40.93  & -76.10  & 0.00  & -3.38  & -0.08  & -0.24  & -38.74  & 3.18560E+01 \\
    mGWO  & 0.20  & 3.55  & 9.04  & 0.21  & -68.28  & -103.82  & 0.00  & -3.38  & -0.08  & -0.24  & -66.05  & 5.61681E+01 \\
    AGWO  & 0.20  & 3.55  & 9.05  & 0.21  & -58.56  & -141.61  & 0.00  & -3.38  & -0.08  & -0.24  & -49.80  & 20.90517405 \\
    SOGWO & 0.20  & 3.51  & 9.04  & 0.21  & -41.39  & -80.33  & 0.00  & -3.38  & -0.08  & -0.24  & -43.23  & 14.06063809 \\
    SCA   & 0.26  & 4.61  & 8.82  & 0.30  & -3230.00  & -6143.49  & -0.04  & -2.68  & -0.13  & -0.24  & -21787.95  & 2.66535E+00 \\
    WOA   & 0.30  & 3.66  & 7.94  & 0.36  & -575.55  & -945.10  & -0.06  & -2.89  & -0.17  & -0.23  & -62619.11  & 3.34546E+01 \\
    \bottomrule
    \end{tabular}}
\end{center}
\end{table}%

\begin{table}[htbp]
\begin{center}
  \caption{Statistical results of five methods for welded beam design problem} \label{res:WBD2}
  \resizebox{4.8in}{15mm}{
      \tabcolsep 12pt 
    \begin{tabular}{ccccc}
    \hline
    Algorithms & Best  & Mean  & Worst & Standard Deviation \\
    \hline
    EBGWO & \textbf{1.72621E+00} & \textbf{1.72871E+00} & \textbf{1.74501E+00} & \textbf{3.45109E-03} \\
    GWO   & 1.72759E+00 & 3.18560E+01 & 2.68650E+02 & 6.32292E+01 \\
    mGWO  & 1.73299E+00 & 5.61681E+01 & 3.36268E+02 & 1.01044E+02 \\
    AGWO  & 1.73197E+00 & 2.09052E+01 & 1.51251E+02 & 3.37442E+01 \\
    SOGWO & 1.72754E+00 & 1.40606E+01 & 1.63283E+02 & 3.19561E+01 \\
    SCA   & 1.95681E+00 & 2.66535E+00 & 4.58405E+00 & 6.41193E-01 \\
    WOA   & 1.80357E+00 & 3.34546E+01 & 4.80804E+02 & 9.07359E+01 \\
    \hline
    \end{tabular}}
\end{center}
\end{table}%

The experimental and statistical results are provided in Tables \ref{res:WBD1} and \ref{res:WBD2}. Compared with the other algorithms, the elite inheritance mechanism of EBGWO improves the search speed and search quality of the algorithm. The EBGWO algorithm can obtain a better solution of the problem with the same number of iterations.

Overall, the EBGWO algorithm has a superior performance compared with other popular algorithms in solving real-world engineering problems.

\section{Conclusion and future works}\label{sec:conclusion}
In this paper, a novel algorithm, i.e., the EBGWO algorithm is proposed to enhance the performance of the GWO algorithm by employing the elite inheritance mechanism and balance search mechanism. The elite inheritance mechanism retains the elite leading wolves from both the last iteration and the current iteration and selects the best three candidate wolves from the candidate pool for population update guidance, which can improve the convergence speed and effect. The position updating method of the original GWO algorithm is calculated by averaging the positions of three wolves. The operation of the balance search mechanism introduces a new operator, $ST$, and devises a new selection method for leading wolves, thereby expanding the search region and facilitating escape from local optima. By conducting a comparative analysis of the EBGWO with the other algorithms, we conclude that the EBGWO algorithm can improve the accuracy and efficiency of optimization, exploration and exploitation abilities. 

The key \textbf{contributions} of this novel algorithm can be summarized as follows:
\begin{itemize}
\item The elite inheritance mechanism of the novel EBGWO algorithm increases the convergence speed and speeds up the efficiency of the algorithm solution. 
\item The balance search mechanism of the novel EBGWO algorithm improves the quality of solutions and enhances the effectiveness of the optimal solution. 
\item The combination of the elite inheritance mechanism and balance search mechanism in the EBGWO algorithm maintains the population diversity, reduces the local optimal stagnation possibility and improves the exploration and exploitation abilities. 
\item Three experiments reveal that the novel EBGWO algorithm has superior performance and can be applied to a set of different real-world design problems.
\end{itemize}

There are several areas for future investigation.
\begin{itemize}
\item Future research could explore modifications to the EBGWO algorithm, such as altering the hierarchy of the wolf population.
\item The capability of the EBGWO algorithm can be improved to handle new types of problems, such as large-scale problems and multi-objective problems.  
\end{itemize}

\section*{Acknowledgment}
The authors thank the financial support from the Foundation of the Jilin Provincial Department of Science and Technology (No.YDZJ202201ZYTS565) and the Foundation of Social Science of Jilin Province, China (No.2022B84).

\bibliographystyle{model5-names}
\biboptions{authoryear}
\bibliography{ref}
\end{document}